\documentclass[10pt,twocolumn,letterpaper]{article}

\usepackage{wacv}              

\usepackage[accsupp]{axessibility} 
\usepackage{graphicx}
\usepackage{amsmath}
\usepackage{amssymb}
\usepackage{booktabs}
\usepackage{xcolor}
\usepackage{subcaption}

\usepackage[pagebackref,breaklinks,colorlinks]{hyperref}

\usepackage[capitalize]{cleveref}
\crefname{section}{Sec.}{Secs.}
\Crefname{section}{Section}{Sections}
\Crefname{table}{Table}{Tables}
\crefname{table}{Tab.}{Tabs.}

\def\wacvPaperID{966} 
\def\confName{WACV}
\def\confYear{2024}

\begin{document}

\title{Neural Echos: Depthwise Convolutional Filters \\ Replicate Biological Receptive Fields}

\author{Zahra Babaiee  \\
TU Vienna, MIT\\
{\tt\small zbabaiee@mit.edu}
\and
Peyman M. Kiasari  \\
University of Waterloo \\
{\tt\small p2mohsen@uwaterloo.ca}
\and
Daniela Rus  \\
MIT \\
{\tt\small rus@mit.edu}
\and
Radu Grosu \\
TU Vienna \\
{\tt\small radu.grosu@tuwien.ac.at} 
}
\maketitle

\begin{abstract}
   In this study, we present evidence suggesting that depthwise convolutional kernels are effectively replicating the structural intricacies of the biological receptive fields observed in the mammalian retina. We provide analytics of trained kernels from various state-of-the-art models substantiating this evidence. Inspired by this intriguing discovery, we propose an initialization scheme that draws inspiration from the biological receptive fields. Experimental analysis of the ImageNet dataset with multiple CNN architectures featuring depthwise convolutions reveals a marked enhancement in the accuracy of the learned model when initialized with biologically derived weights. This underlies the potential for biologically inspired computational models to further our understanding of vision processing systems and to improve the efficacy of convolutional networks.
\end{abstract}


\section{Introduction}
\label{sec:intro}

Convolutional Neural Networks (CNNs)~\cite{doi:10.1162/neco.1989.1.4.541}, a mainstay of modern artificial intelligence (AI), owe their fundamental design principles to insights drawn from neuroscience (NS)~\cite{mack2013principles}, particularly our understanding of receptive fields. A receptive field is the specific region of sensory space eliciting a response from a neuron when stimulated~\cite{hartline_1940,mack2013principles}. The concept is deeply ingrained in the architecture of the mammalian visual system, starting from the retina. CNNs mimic this structure through their use of 'kernels' capable of responding to a specific part of the image. This convolution process mirrors the hierarchal, spatially invariant nature of biological vision systems, underscoring the deep connections between the fields of NS and AI.

\begin{figure}[t]
  \centering
   \includegraphics[width=0.9\linewidth]{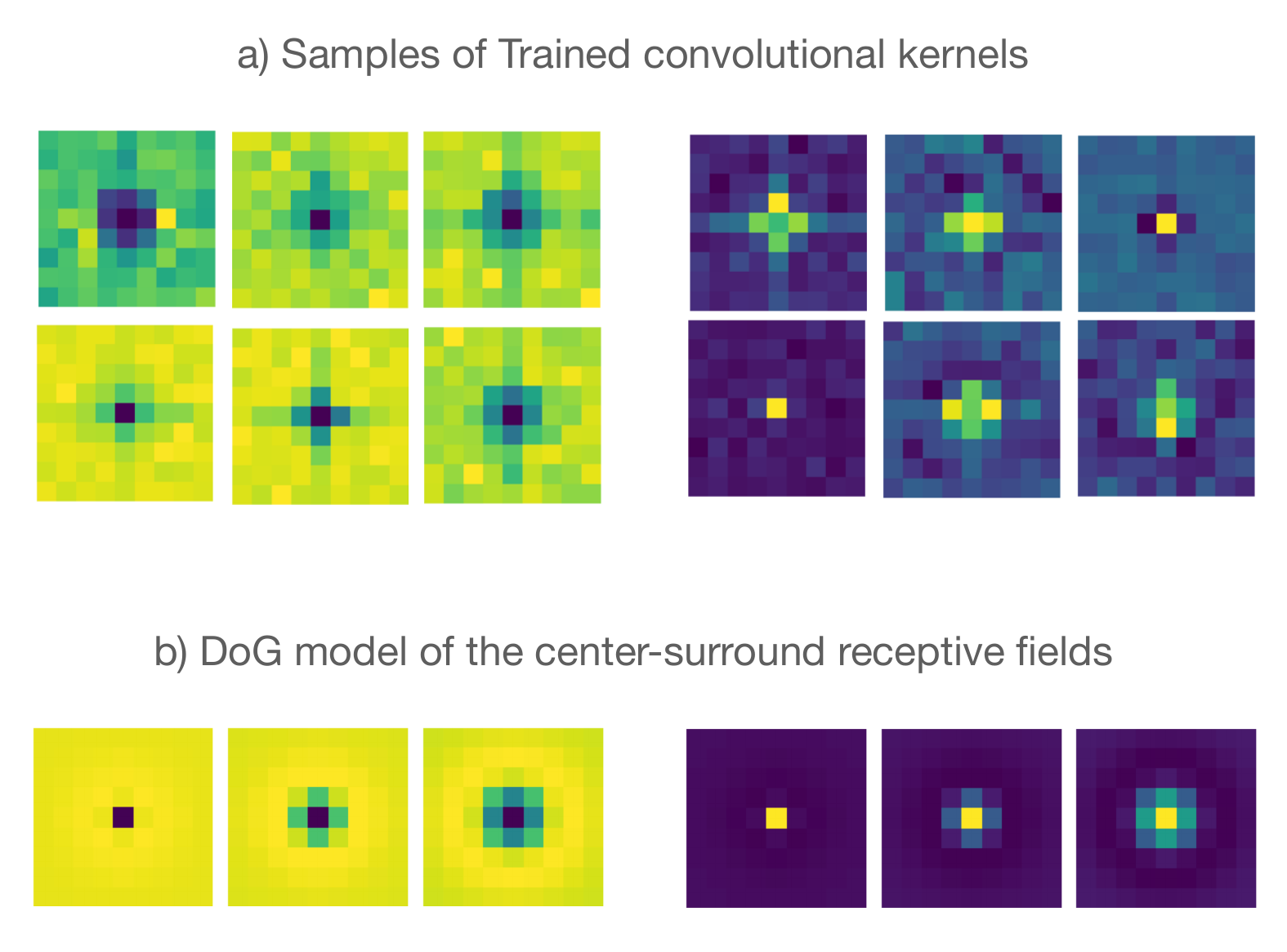}

   \caption{a) Depthwise Convolutional kernels trained on ImageNet dataset, and b) the DoG model of the biological center-surround receptive fields with different center-to-surround ratios, and with excitatory center (right) and inhibitory center (left), respectively. Artificial kernels mimic biological center-surround patterns.}
   \label{fig:onecol}
   \vspace{-2.5ex}
\end{figure}

The realm of convolutional neural networks (CNNs) has witnessed remarkable evolutionary phases since its inception. Initial architectures, such as AlexNet~\cite{alexnet}, introduced in 2012, focused on varying kernel sizes to capture image features. As the field matured, architectures like VGG networks~\cite{DBLP:journals/corr/SimonyanZ14a} and Residual Networks~\cite{he2016residual} standardized the use of 3x3 kernels, optimizing for efficiency and training speed. However, a pivotal shift emerged with the introduction and popularization of depthwise convolutions.

Depthwise convolutions introduced a novel approach to feature extraction, where each input channel is individually convolved with its own filter, as opposed to standard convolutions that aggregate information across multiple channels. This technique, exemplified by architectures like MobileNet with its 3x3 depthwise convolutions, offers significant reductions in computational overhead without markedly sacrificing model accuracy. The advent of vision transformers and their patch-centric designs~\cite{imageWorth16Words} further accentuated the exploration into depthwise convolution behaviors with larger kernel sizes, reinforcing their unique ability to manifest structured patterns. 

\begin{figure}
  \centering
   \includegraphics[width=0.95\linewidth]{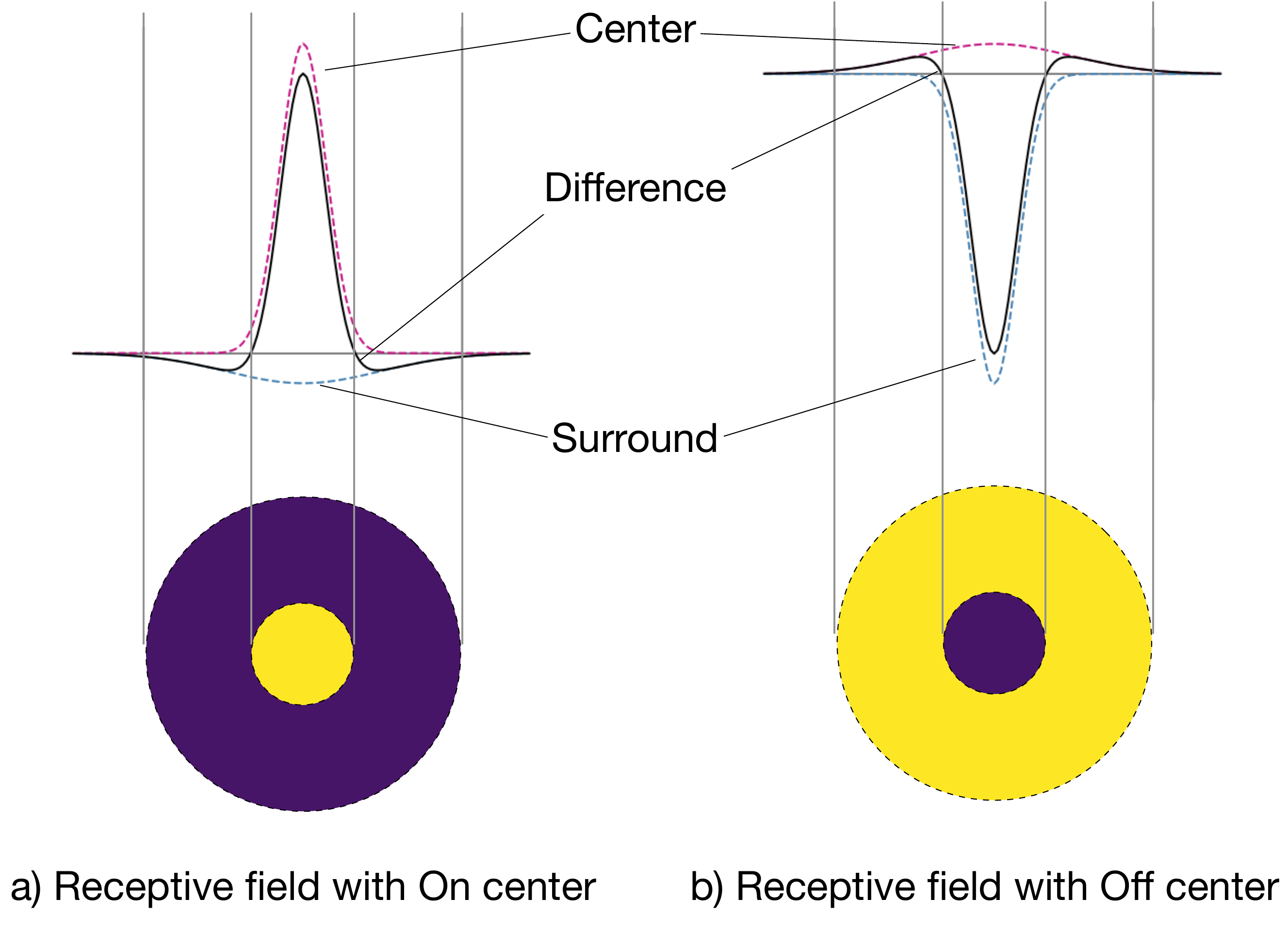}

   \caption{A ``difference-of-Gaussians" is used to model a neuron's sensitivity to light at various positions on the retina. This model comprises two Gaussian functions - a narrow, positive one, representing the stimulatory center, and a wide, negative one, indicating the suppressive surround, for the neurons with an excitatory center, and the other way around for the ones with an inhibitory center.}
   \label{fig:DoG}
   \vspace{-2.5ex}
\end{figure}



The cornerstone of visual processing in numerous retinal cell types, including the intricate network of ganglion neurons, is the principle of center-surround antagonism, a mechanism established in the receptive field of neurons, as early as in the retina~\cite{doi:10.1152/jn.1953.16.1.37,mack2013principles, enroth-cugell_pinto_1972}. This mechanism stems from lateral inhibitory connections and is perpetuated by neurons in higher visual processing centers, namely the lateral geniculate nucleus and the visual cortex~\cite{hubelWiesel1968}.

The center-surround antagonism plays a vital role in the primate visual system, assisting in complex tasks such as edge detection, figure-background segregation, depth, and object perception, that remain consistent across various visual cues. Importantly, this architecture has two key configurations: excitatory- and inhibitory-center receptive fields, respectively~\cite{Zaghloul2645,SHAPLEY1986229}. In the former configuration, ganglion cells are excited by light falling on the center of the receptive field and inhibited by light falling on the surrounding area. Conversely, in the latter configuration, cells are inhibited by light at the center and excited by light in the surrounding area. This design enhances contrast and aids in edge detection~\cite{enroth-cugell_pinto_1972,mack2013principles}.

Classical NS models frequently employ receptive fields featuring center-surround antagonism, typically realized through a Difference of Gaussians (DoG) function, which creates an excitatory peak at the receptive field's center counterbalanced by an inhibitory surround\cite{jacobsen2016structured, fukushima_2003}. 

 In our investigations, we unearthed a remarkable parallel between the trained kernels of depthwise convolutions in various models and biological receptive fields: a significant quantity of them echoed the center-surround pattern seen in biological receptive fields. Intriguingly, such patterns were exclusively observed in depthwise convolutions, eluding their regular convolution counterparts. In Figure~\ref{fig:onecol} we provide a comparative demonstration of the trained depthwise kernels and the NS-based model of center-surround kernels, highlighting their noteworthy similarities. This discovery underscores not only the computational advantages of depthwise convolutions but also their potential to mirror biologically-inspired patterns, reaffirming the value of rethinking standard convolution operations in modern deep-learning paradigms.

Taking cues from these resemblances, we suggest a center-surround initialization procedure for depthwise convolutional kernels. Our experiments on ImageNet dataset with different models revealed that networks when initialized using our biologically-inspired methodology, display a marked increase in accuracy. Specifically, models initialized by our method gain up to more than two percent accuracy on the ImageNet dataset. Despite these notable improvements, the primary purpose of this paper is not solely to underscore performance enhancements. Rather, we aim to emphasize the intriguing discovery that artificial kernels emulate their biological counterparts without explicit supervision. Our results demonstrate the significant potential of biologically inspired computational models in enriching our comprehension of vision processing systems and enhancing the performance of artificial neural networks.


\begin{figure}[t]
  \centering
   \includegraphics[width=0.7\linewidth]{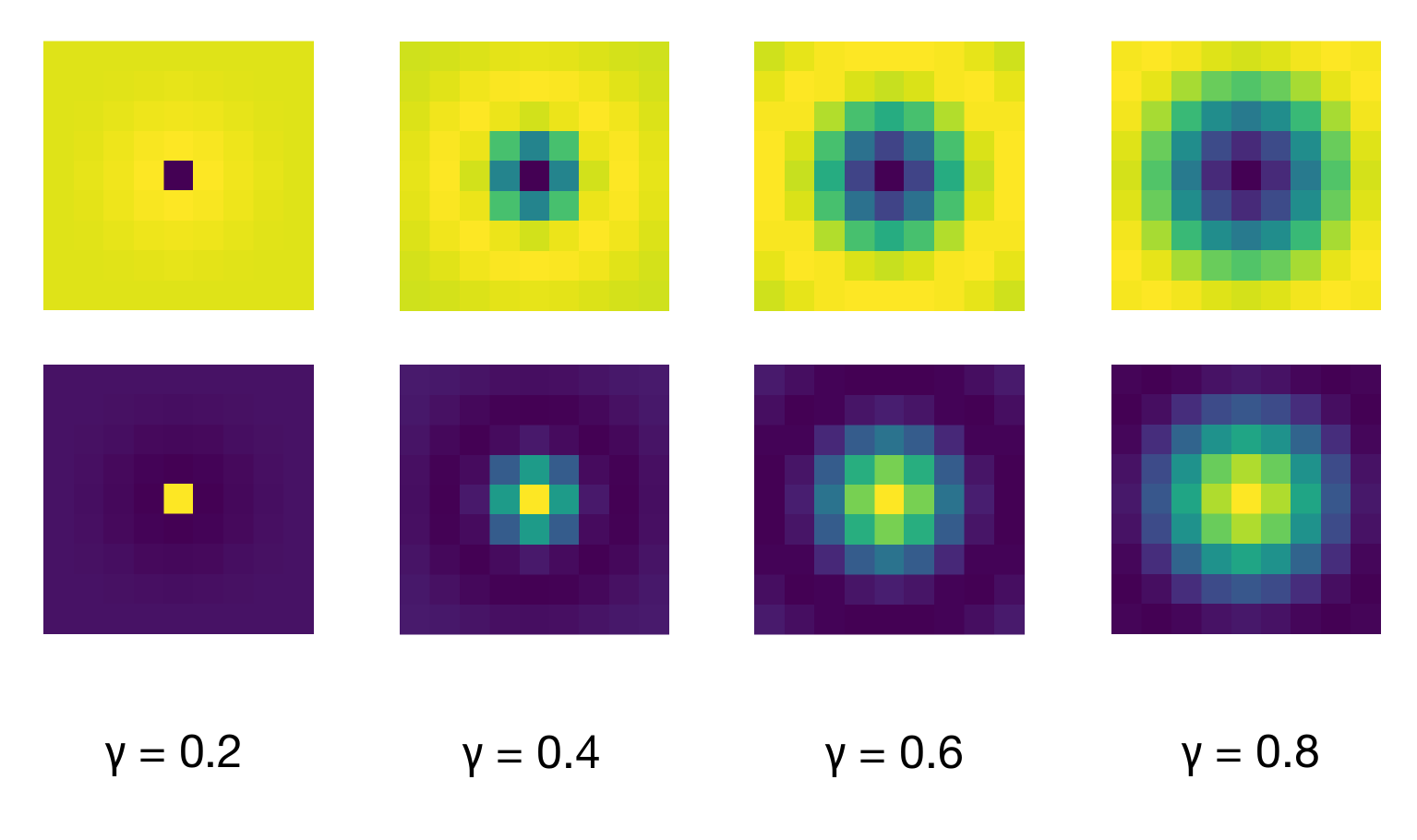}

   \caption{Size 9 DoG kernels with inhibitory (top) and excitatory (buttom) centers, with different ratios of the center-surround radiuses ($\gamma$).}
   \label{fig:kernel_ratios}
   \vspace{-2.5ex}
\end{figure}

\section{Related Work}

\begin{figure*}[t]
  \centering
  
  \begin{subfigure}{0.24\linewidth}
    \includegraphics[width=\linewidth]{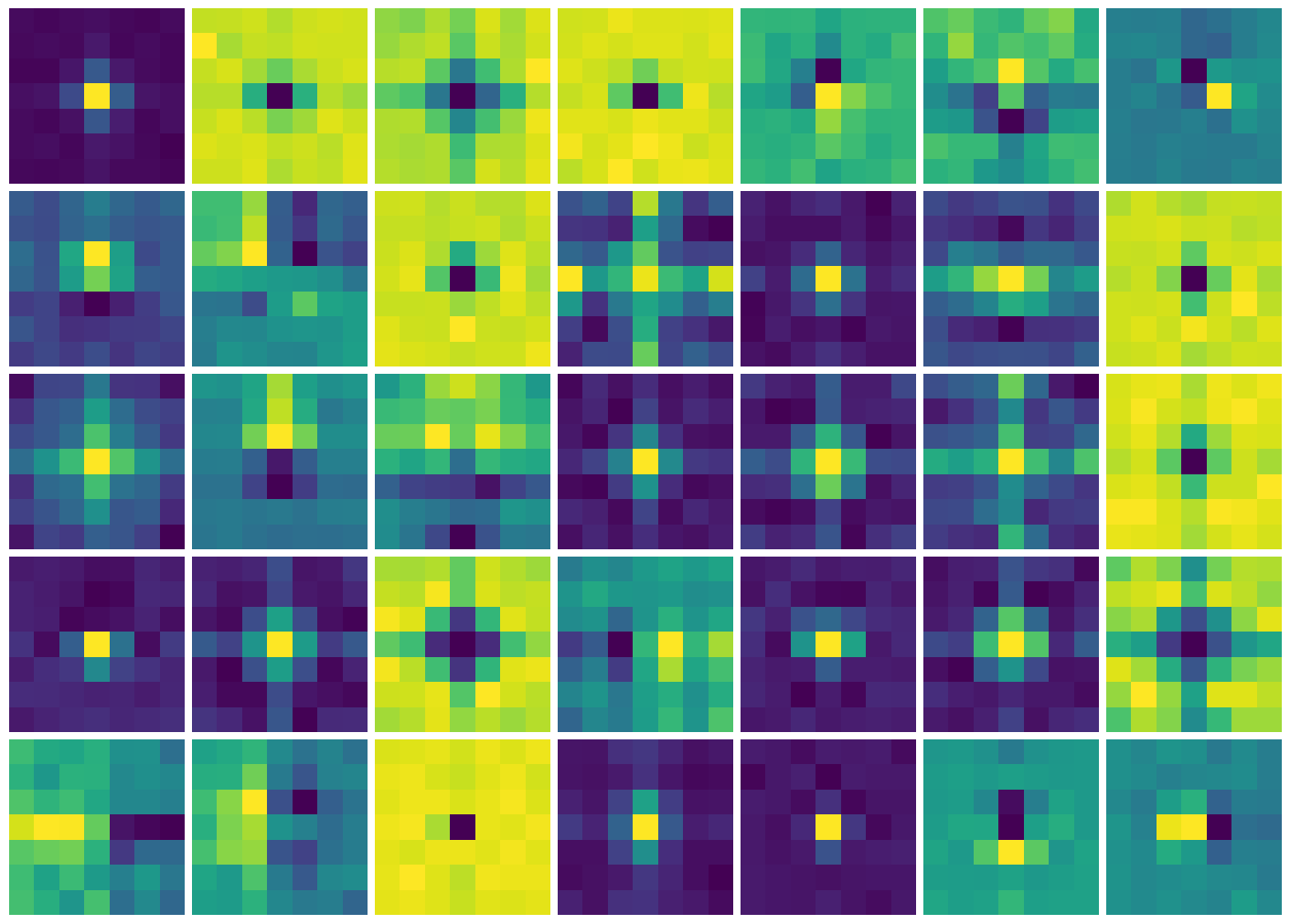}
    \caption{ConvNeXt small, kernel 7×7}
    \includegraphics[width=\linewidth]{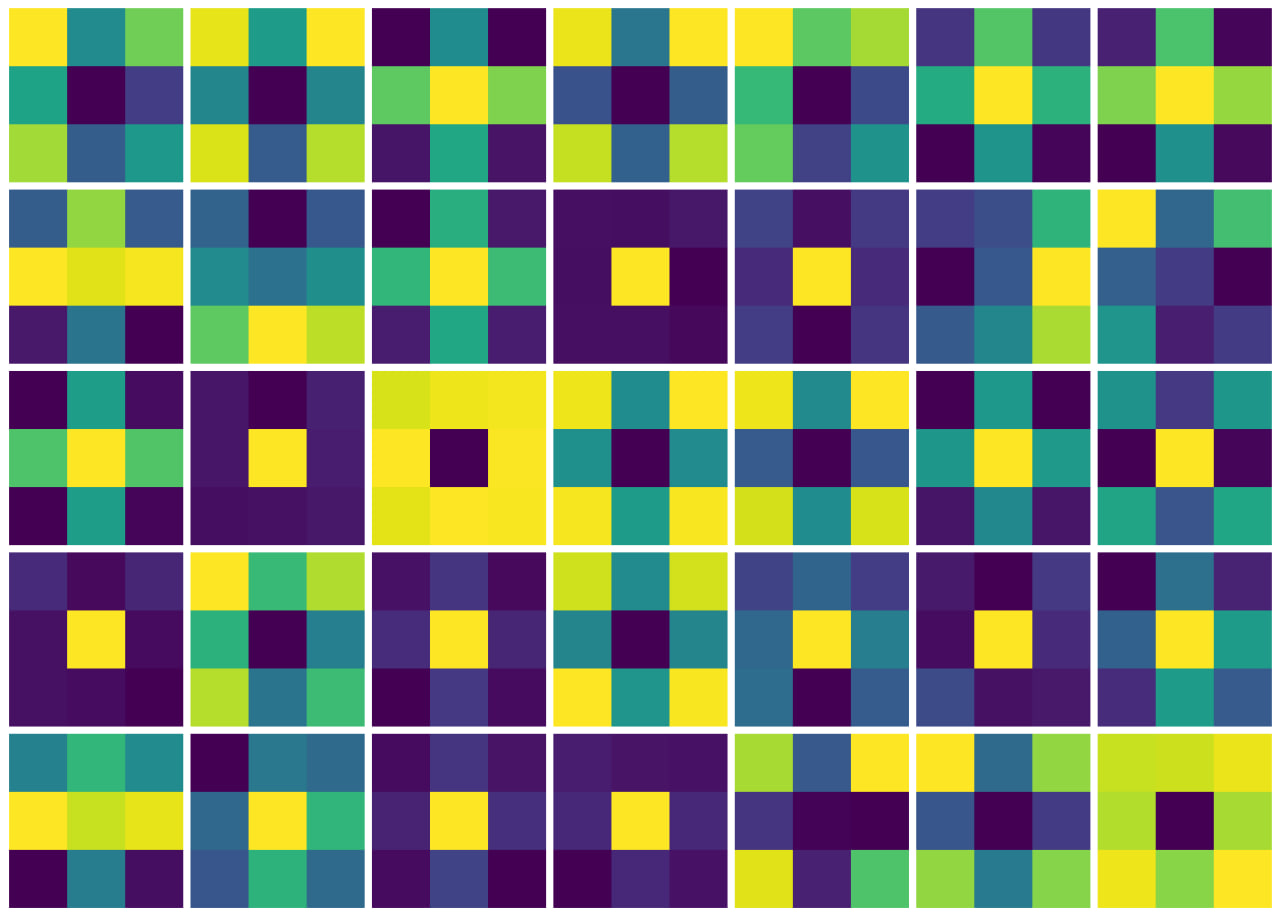}
    \caption{EfficientNet , kernel 3×3}
  \end{subfigure}
  \hfill
  \begin{subfigure}{0.24\linewidth}
    \includegraphics[width=\linewidth]{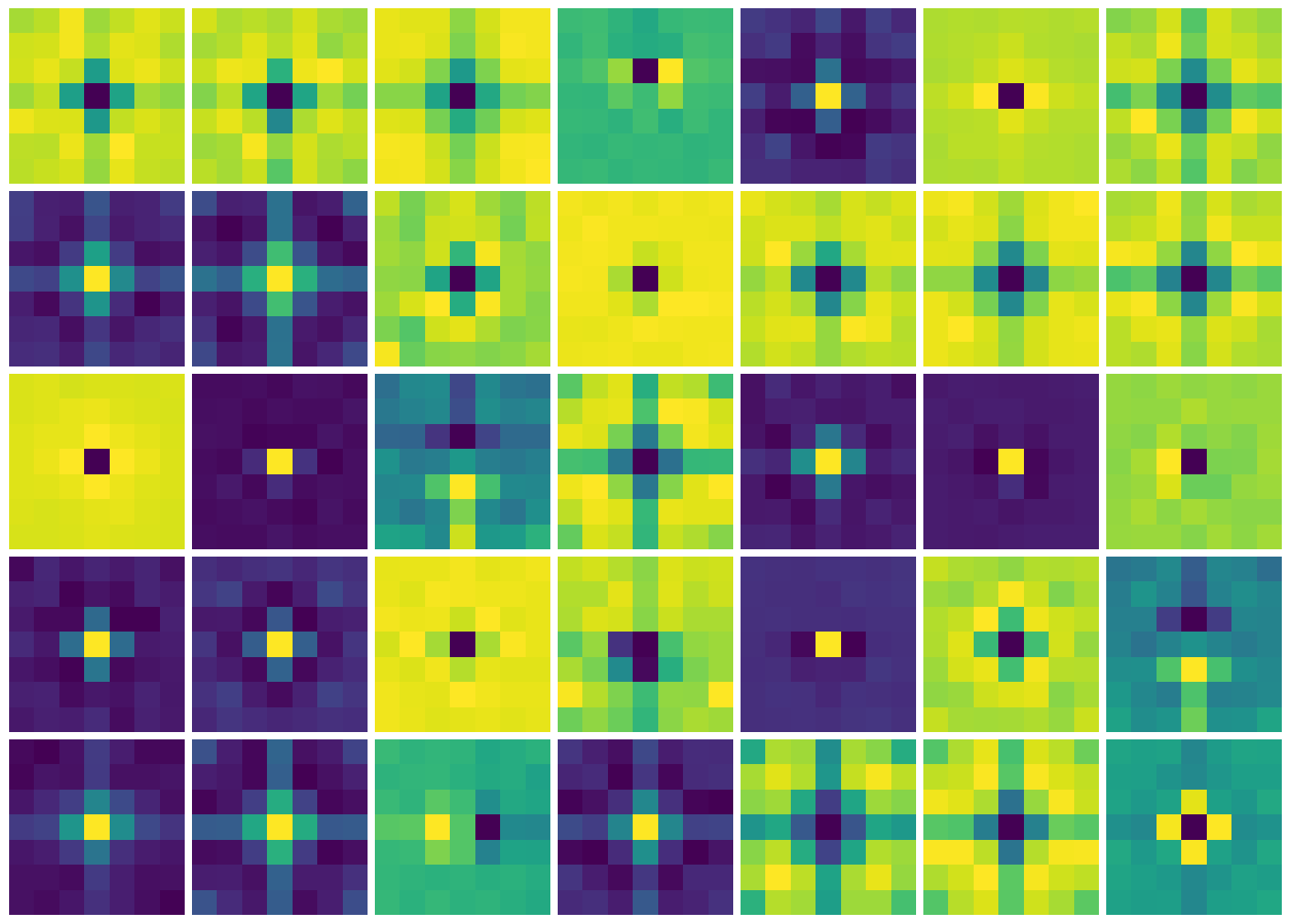}
    \caption{ConvNeXtV2 tiny, kernel 7×7}
    \includegraphics[width=\linewidth]{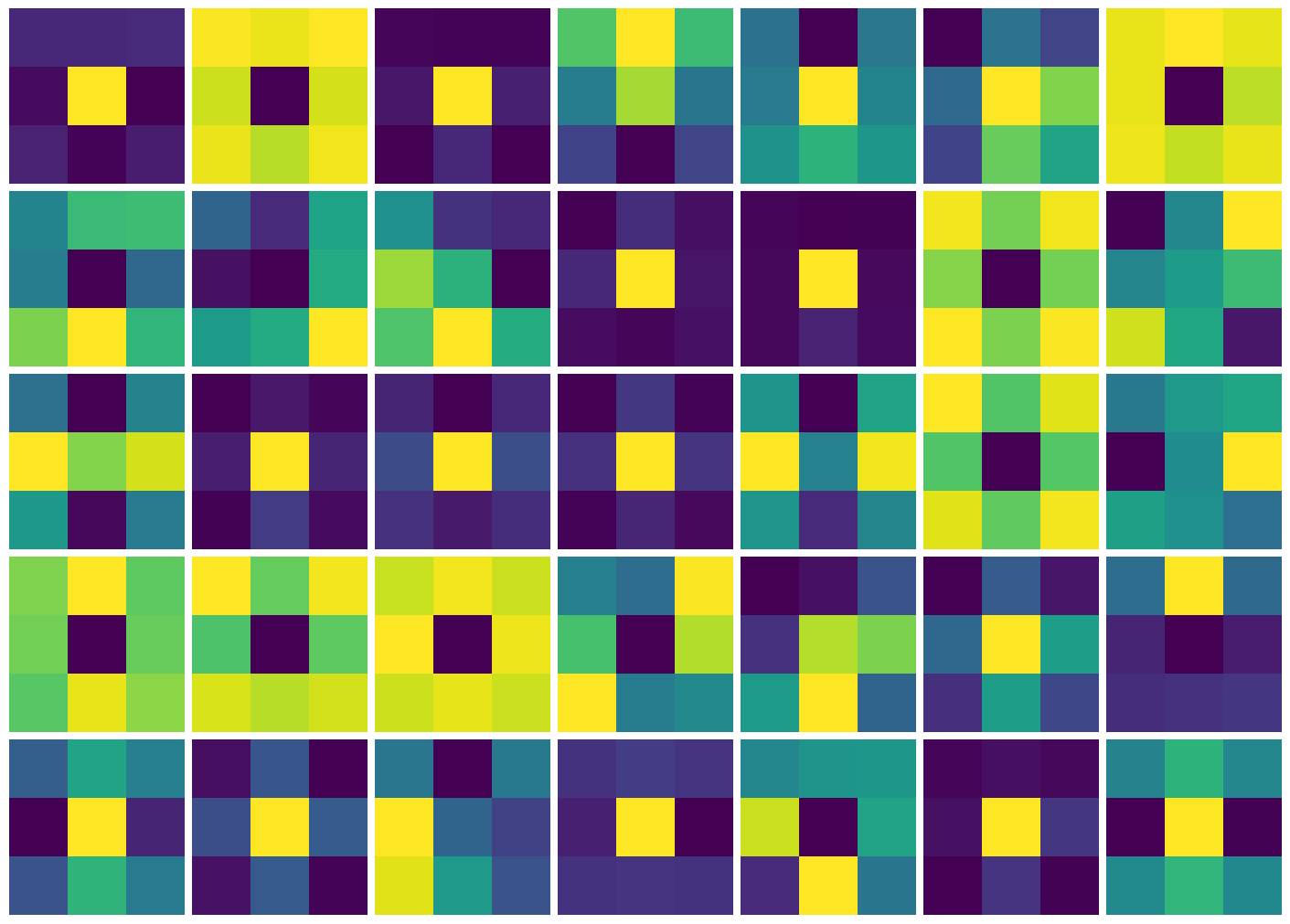}
    \caption{MobileNetV3, kernel 3×3}
  \end{subfigure}
  \hfill
  \begin{subfigure}{0.24\linewidth}
    \includegraphics[width=\linewidth]{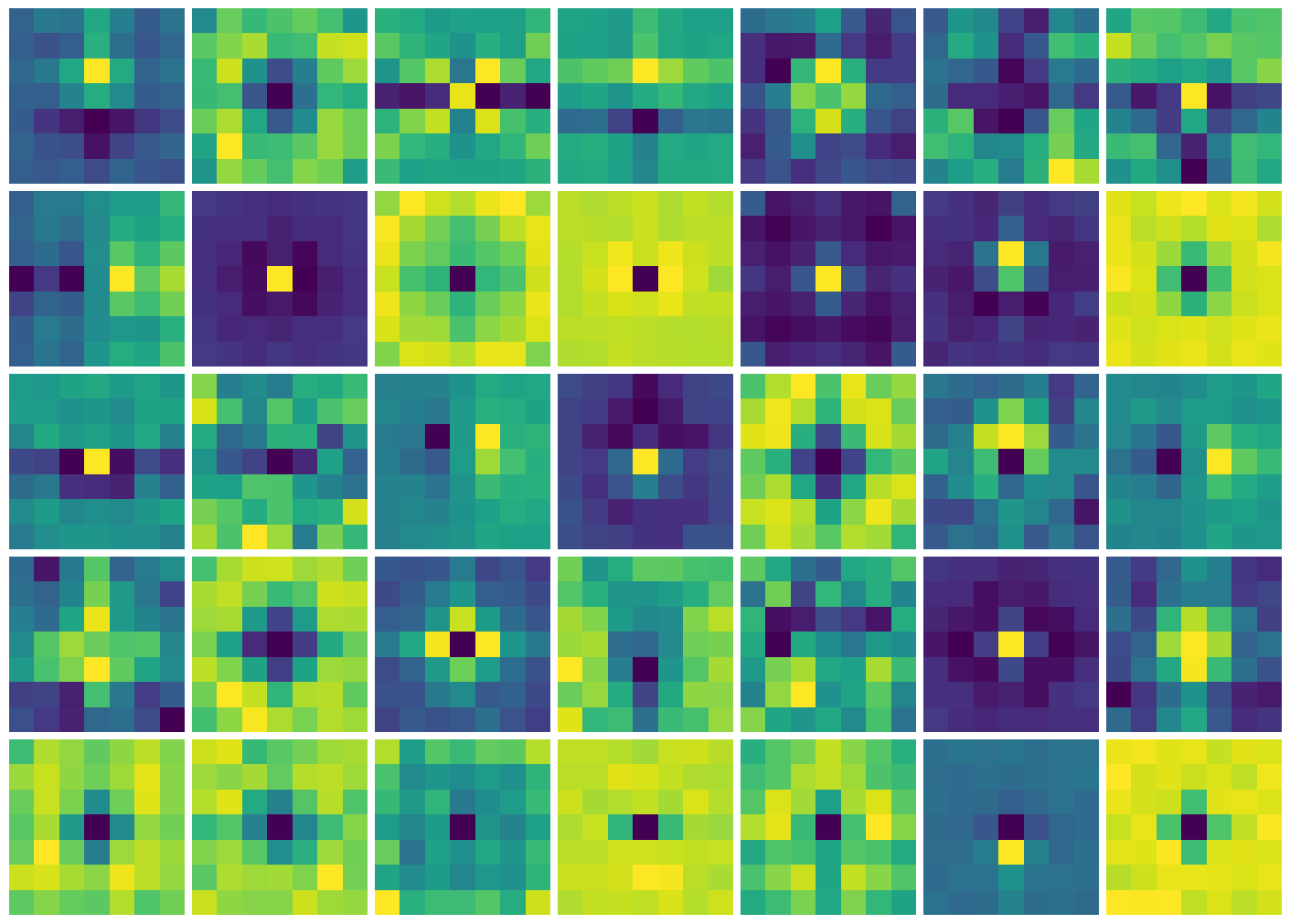}
    \caption{Hornet small, kernel 7×7}
    \includegraphics[width=\linewidth]{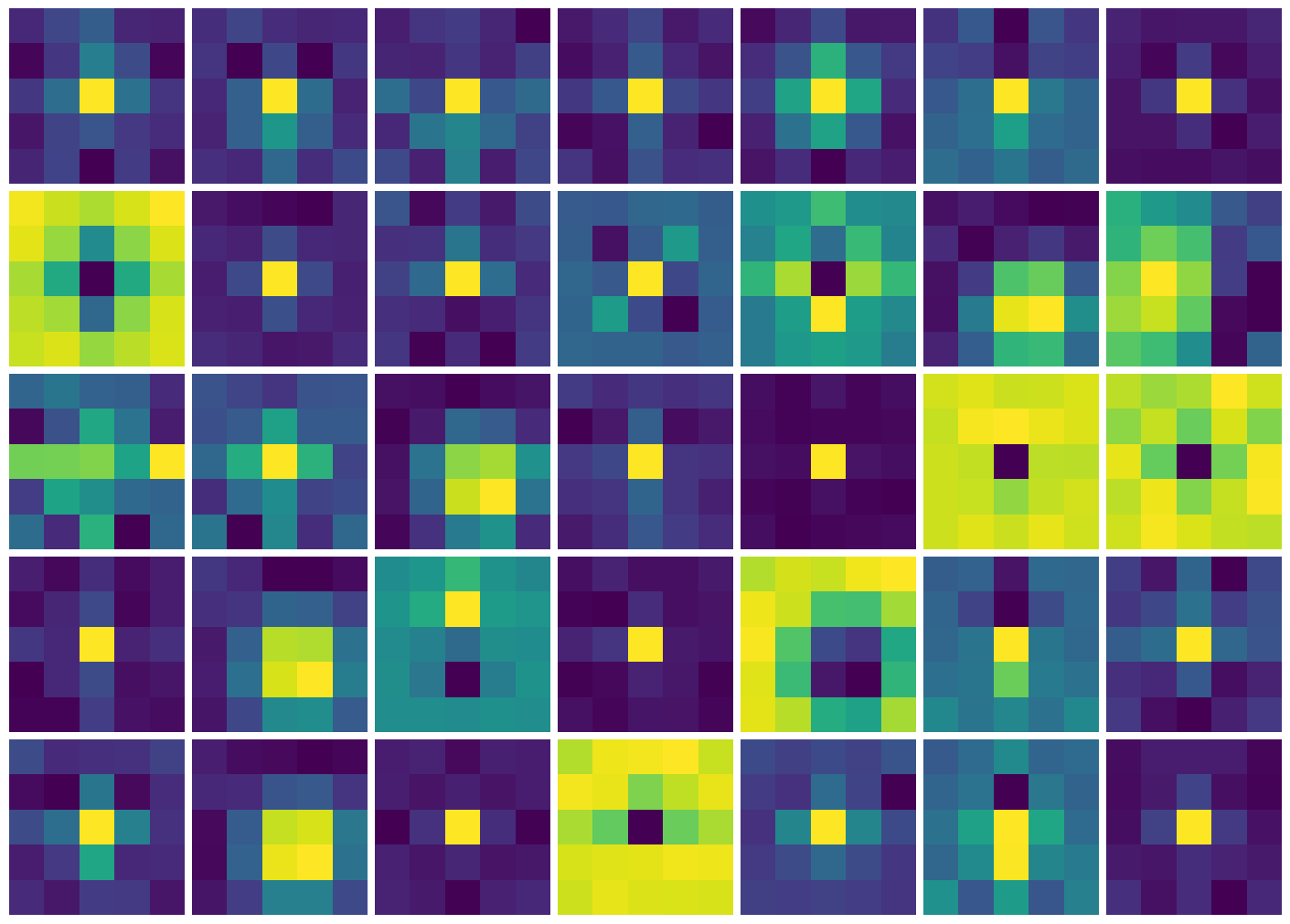}
    \caption{MobileNetV3, kernel 5×5}
  \end{subfigure}
  \hfill
  \begin{subfigure}{0.24\linewidth}
    \includegraphics[width=\linewidth]{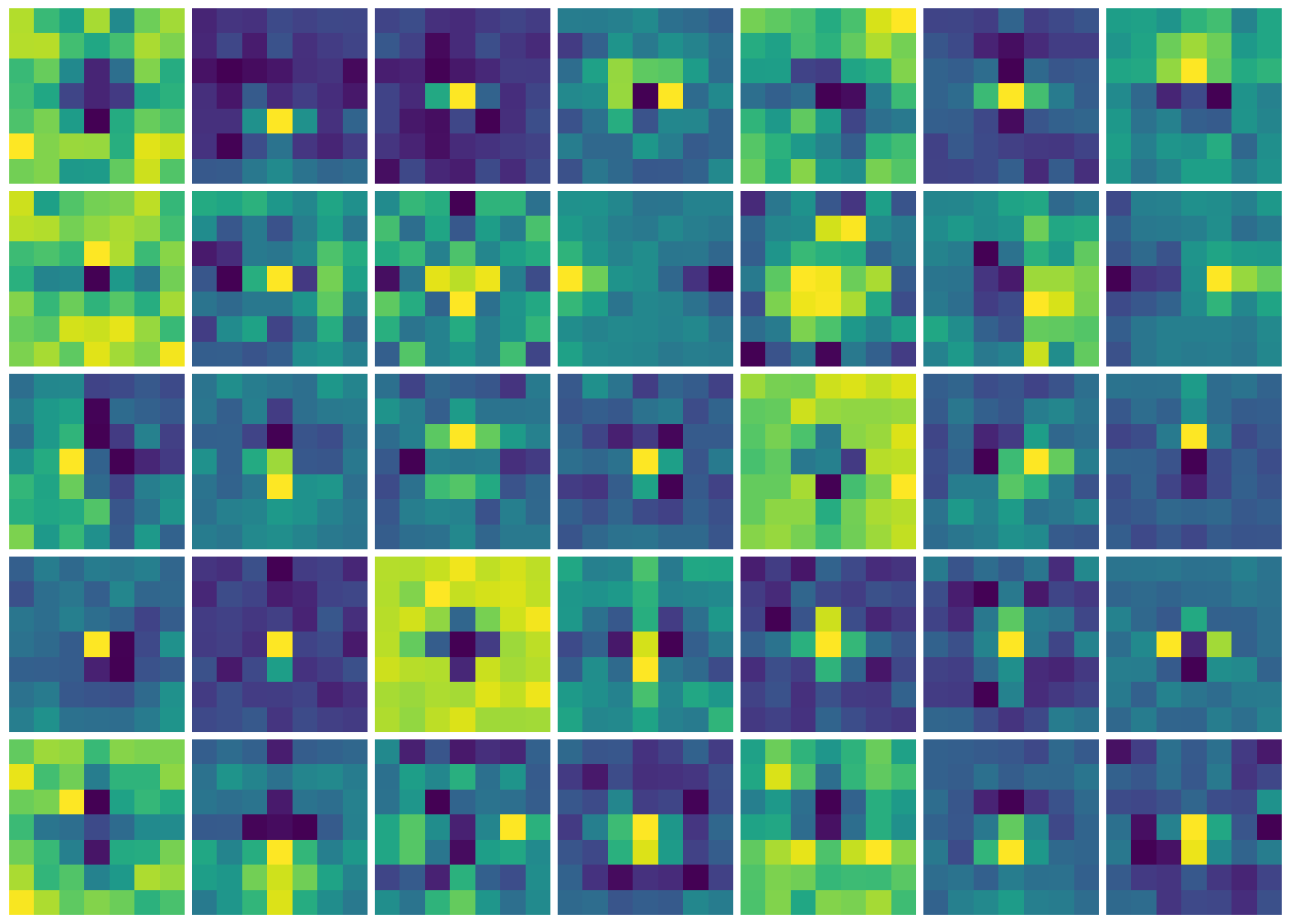}
    \caption{ConvMixer-512, kernel 7×7}
    \includegraphics[width=\linewidth]{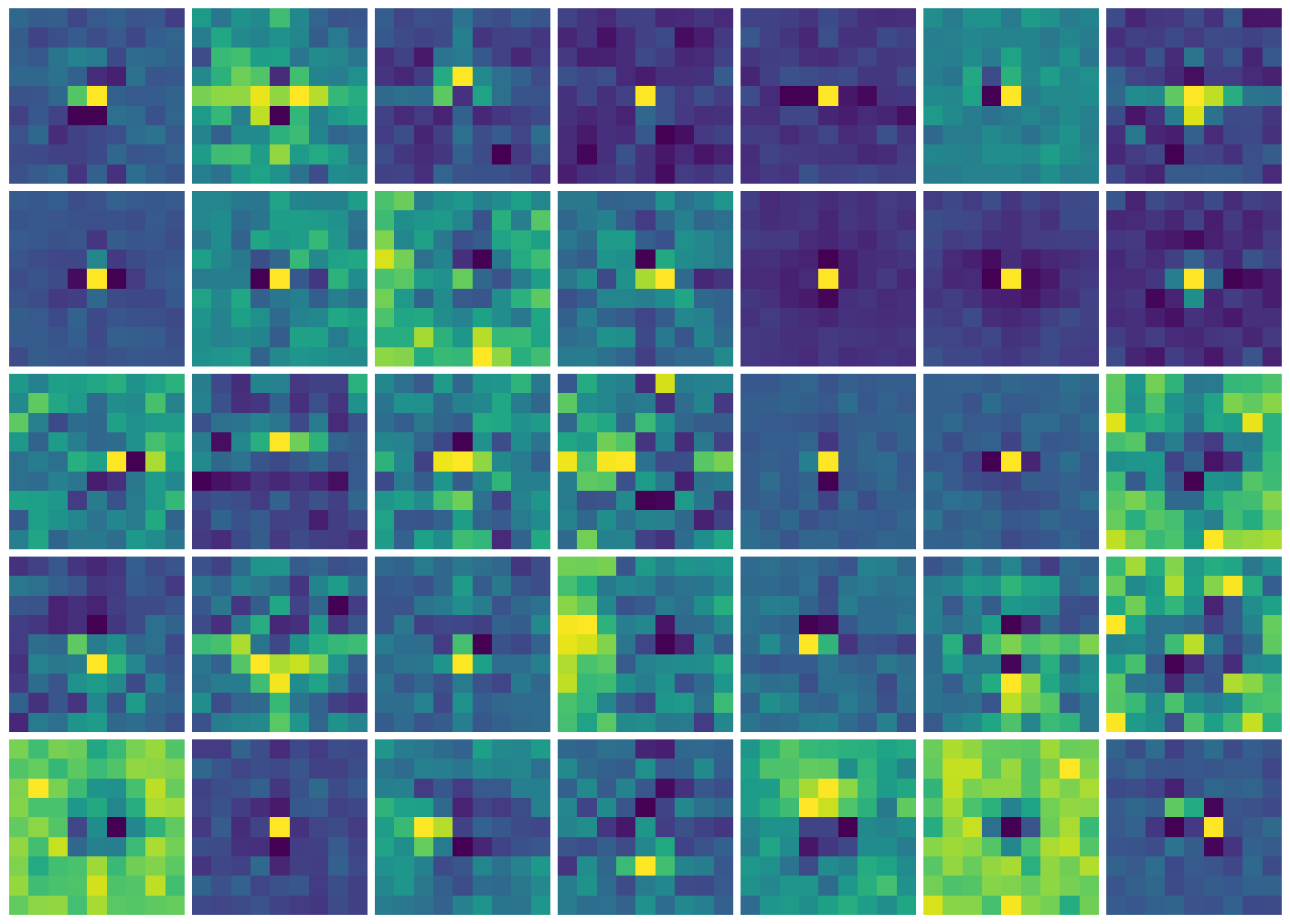}
    \caption{ConvMixer-1024, kernel 9×9}
  \end{subfigure}
  
  \caption{Random samples from depth-wise convolutions of various models with different kernel sizes, trained on the ImageNet dataset. Trained kernels show considerable repeating patterns, many of them featuring a center-surround structure.}
  \label{fig:Mixed_kernels}
  \vspace{-2.5ex}
\end{figure*}

\textbf{Depth-wise Convolutions.}
The evolution of convolutional neural networks has been marked by the introduction and adaptation of diverse convolution operations. Notably, depthwise convolutions, where each input channel is convolved with its distinct filter, have gained traction. With the recent surge of modern enhanced CNN architectures, especially in the wake of the transformative impact of vision transformers, many models are now favoring depthwise convolutions with large kernels over traditional regular convolutions. Depthwise convolutions gained prominence with the introduction of the MobileNet architecture~\cite{mobilenets}, which showcased their efficacy in crafting lightweight models tailored for mobile and embedded vision applications. With the resurgence of the modern CNN architectures after the introduction of vision transformers, many models use depthwise convolutions in their blocks~\cite{convnext, ConvNeXtV2, hornet, convmixer}.

\vspace*{0.5ex}\textbf{Bio-inspired Models.} A considerable volume of research has aimed to incorporate insights from NS into computer vision systems~\cite{kim2016convolutional,zoumpourlis2017non,laskar2018correspondence}. Initial vision models were significantly influenced by NS and psychology. In recent times, there have been substantial advances in both NS and AI, especially in computer vision. However, the majority of contemporary networks, are only loosely based on the visual system, and cross-fertilization between the two fields is less frequent as in the early days of AI. This is in spite of the fact that NS continues to be a vital source of innovative ideas that fuel advancements in AI ~\cite{hassibis, HASSABIS2017245}.

The development of AI models that closely resemble their biological counterparts and that incorporate advances in NS offers two primary advantages. Firstly, NS can be a fertile source of inspiration for designing new models and enhancing existing ones. This holds true both in isolation and in tandem with the computational and mathematical advancements underpinning new models. Secondly, NS can offer validation for existing models and methodologies in the AI domain. An example of this is the residual connections found in pyramidal cells within the cerebral cortex. These connections enable input from layer I to reach cortical layer VI neurons, bypassing intermediary layers~\cite{10.3389/fnana.2010.00013, HASSABIS2017245}.

\vspace*{0.5ex}\textbf{Center-Surround Receptive Fields.}
The Neocognitron model, proposed by Fukushima, holds a key position in the history of neural networks and machine learning, as one of the earliest examples of a CNN. Inspired by the pioneering work of Hubel and Wiesel on the visual cortex of cats~\cite{hubelWiesel1968}, the Neocognitron model was designed to mimic the hierarchical structure of the visual system in mammals. It included a contrast-extracting preprocessing layer, reminiscent of the On-Off ganglion neurons, as well as inhibitory surround connections, mirroring the surround modulation observed in the visual cortex~\cite{fukushima_2003}. More recent studies have tried to incorporate center-surround receptive fields into CNNs by integrating convolutional layers equipped with fixed kernels into the input feature maps. Evidence indicates that this modification enhances the network's performance and resilience, particularly with respect to variations in lighting conditions and input noise~\cite{pmlr-v139-babaiee21a, NEURIPS2019_c535e3a7}.

\begin{figure*}[t]
  \centering
  
  \begin{subfigure}{0.24\linewidth}
    \includegraphics[width=0.3227\linewidth]{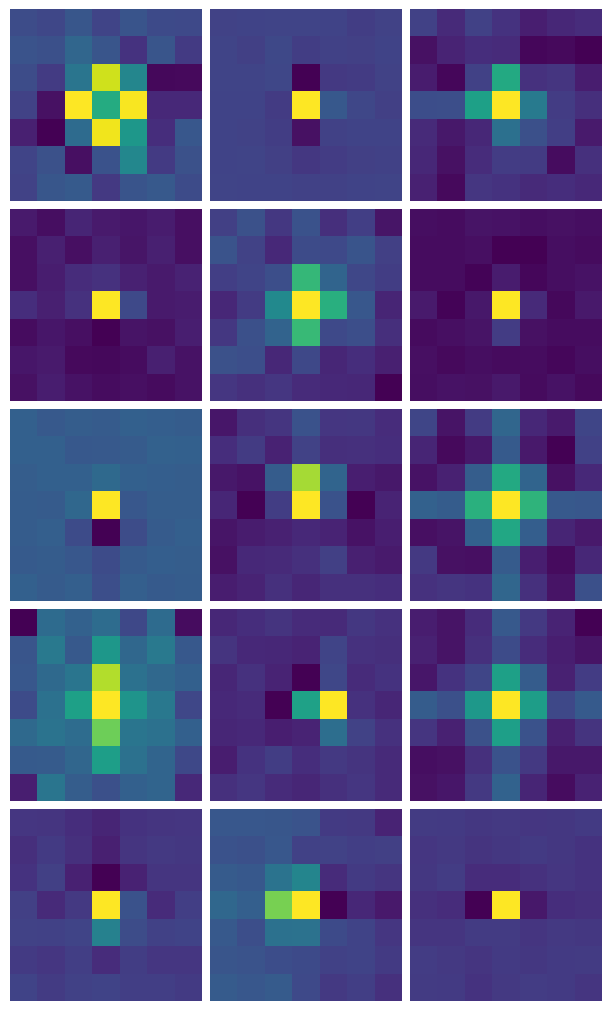}
    \includegraphics[width=0.3227\linewidth]{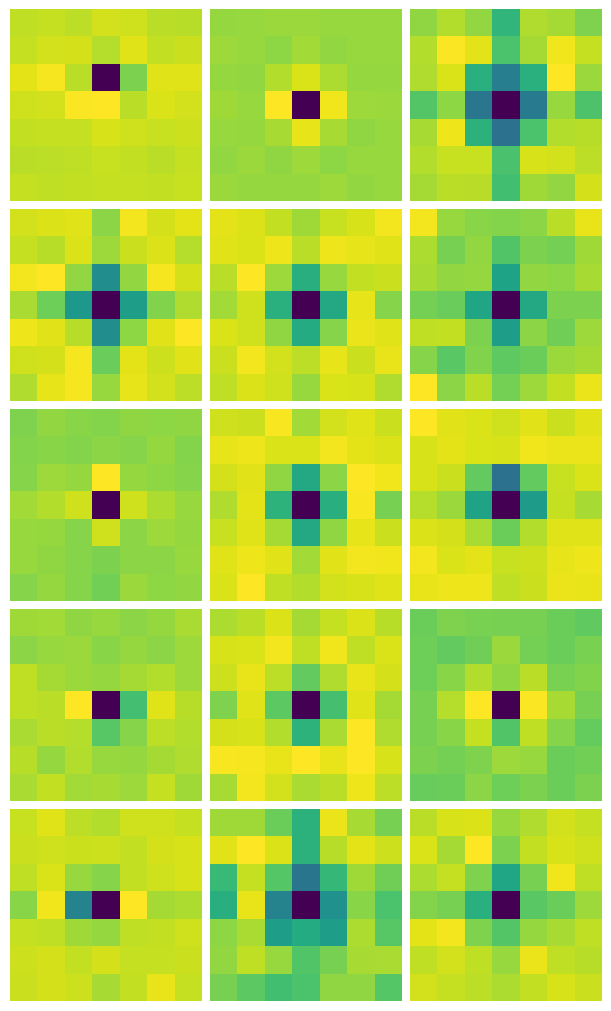}
    \includegraphics[width=0.3227\linewidth]{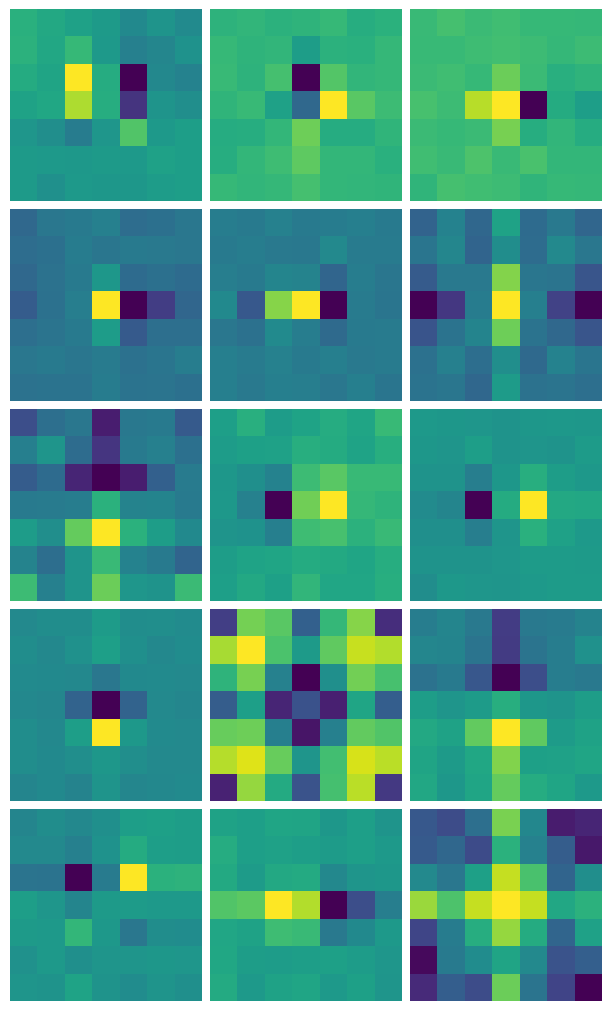}
    \includegraphics[width=0.3227\linewidth]{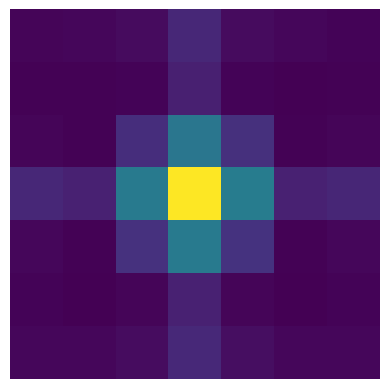}
    \includegraphics[width=0.3227\linewidth]{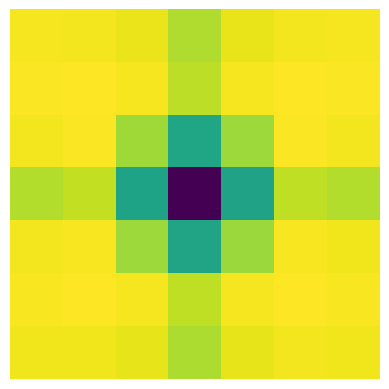}
    \includegraphics[width=0.3227\linewidth]{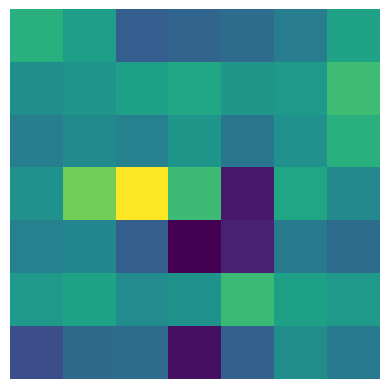}
    \caption{ConvNeXt small, kernel 7×7}
  \end{subfigure}
  \hfill
  \begin{subfigure}{0.24\linewidth}
    \includegraphics[width=0.3227\linewidth]{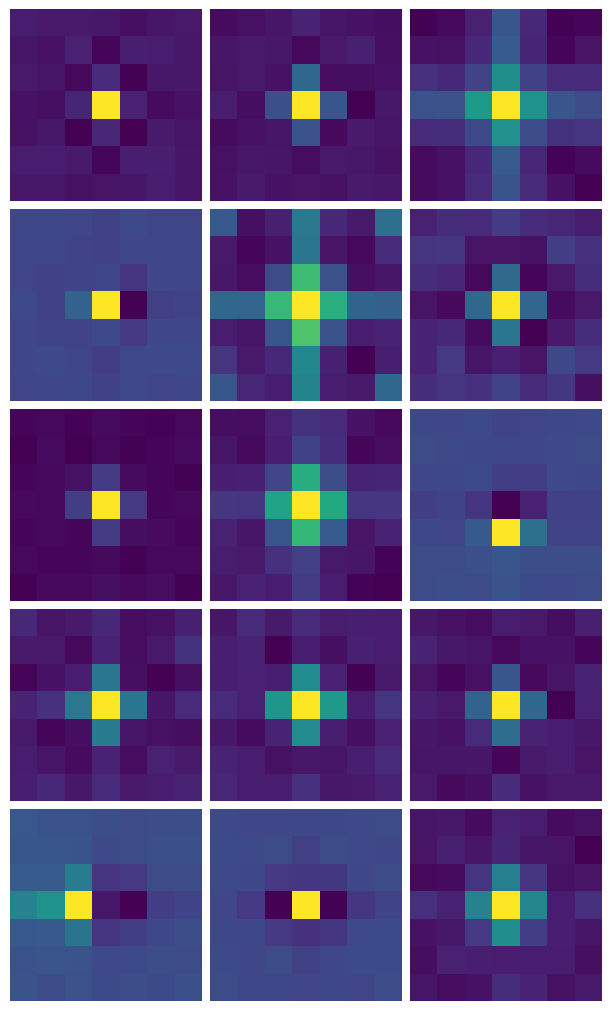}
    \includegraphics[width=0.3227\linewidth]{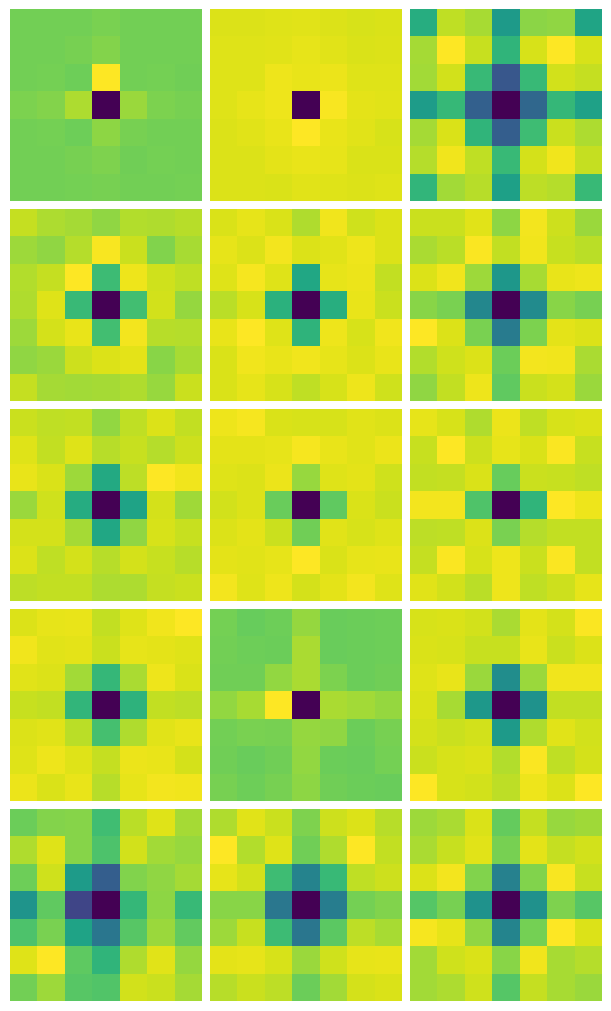}
    \includegraphics[width=0.3227\linewidth]{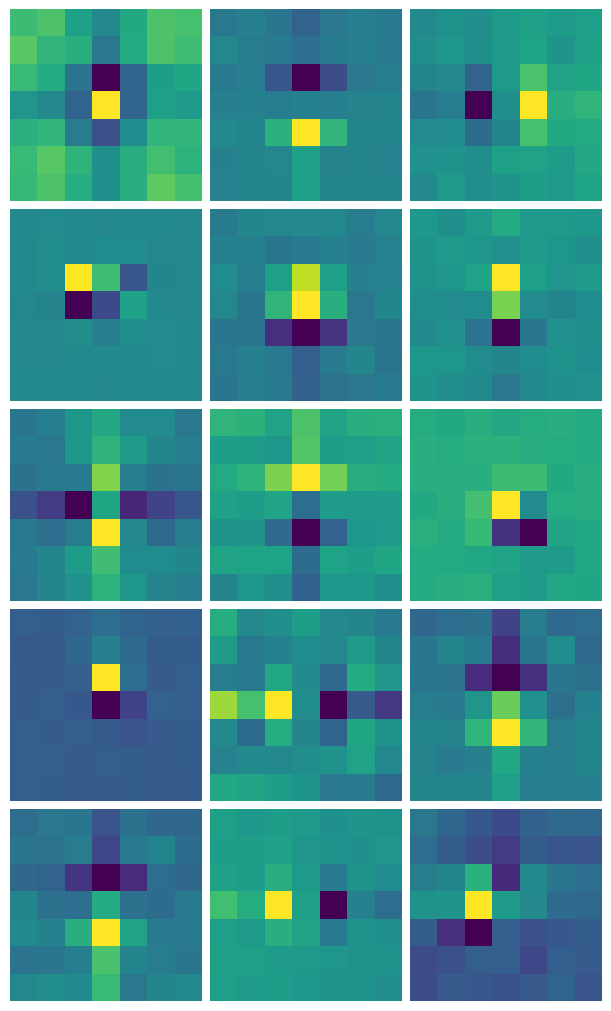}
    \includegraphics[width=0.3227\linewidth]{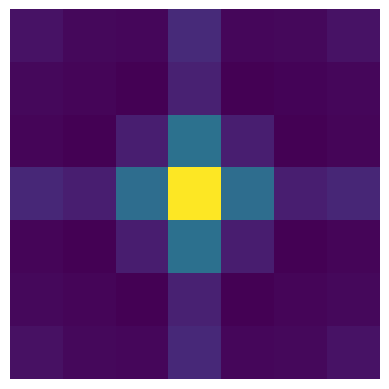}
    \includegraphics[width=0.3227\linewidth]{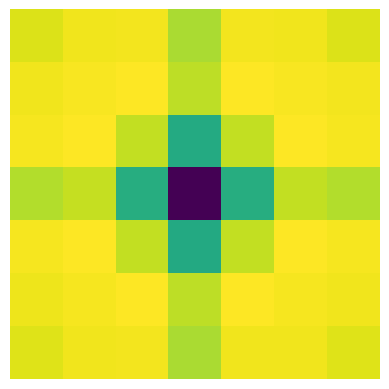}
    \includegraphics[width=0.3227\linewidth]{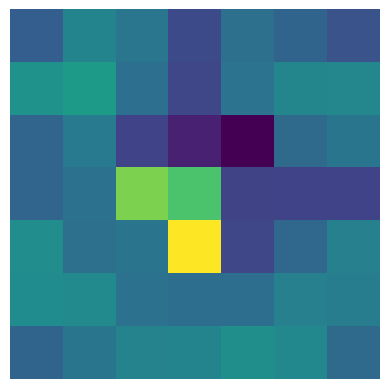}
    \caption{ConvNeXtV2 tiny, kernel 7×7}
  \end{subfigure}
  \hfill
  \begin{subfigure}{0.24\linewidth}
    \includegraphics[width=0.3227\linewidth]{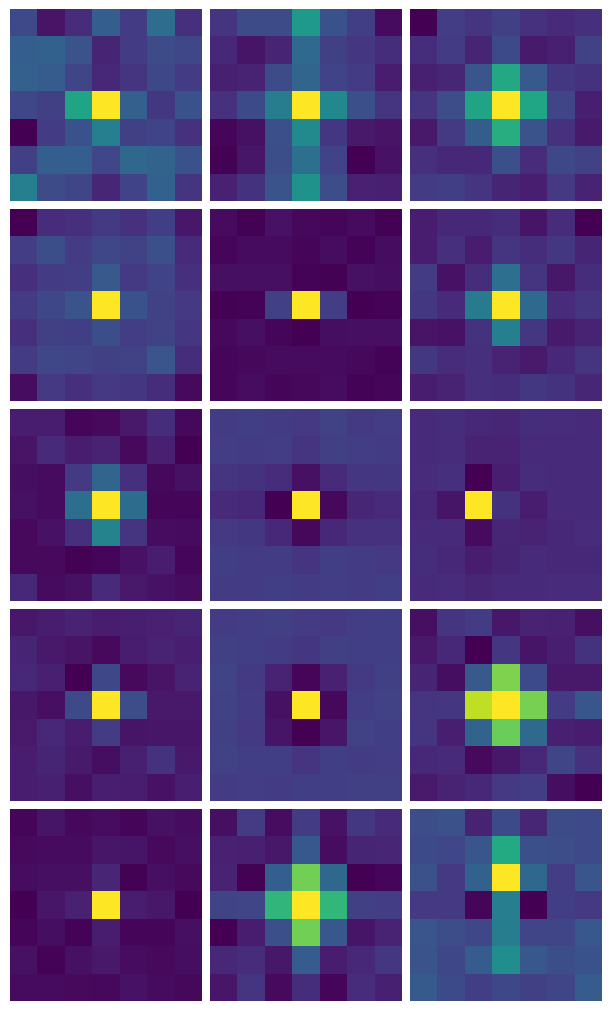}
    \includegraphics[width=0.3227\linewidth]{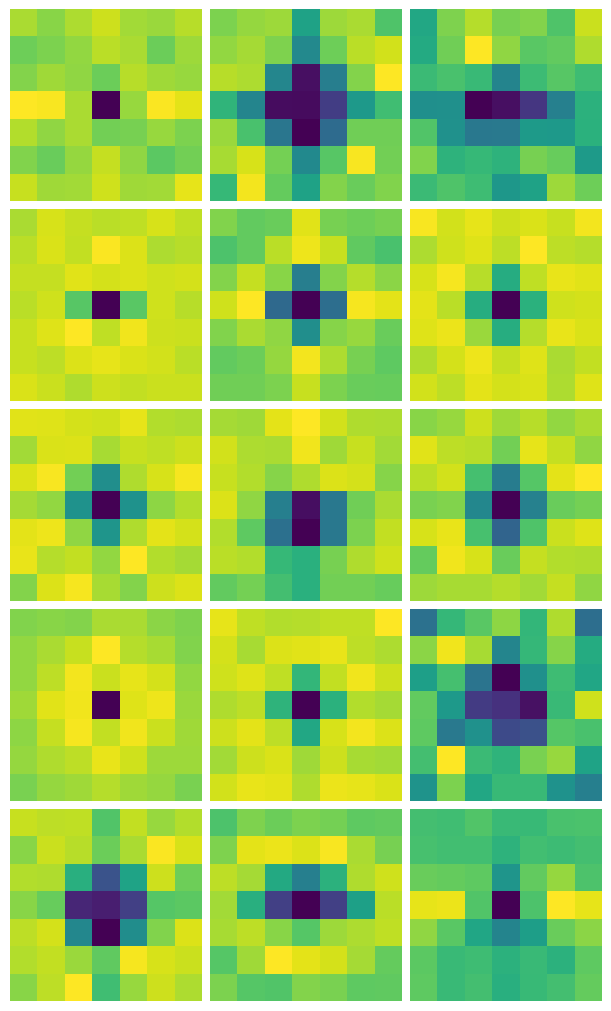}
    \includegraphics[width=0.3227\linewidth]{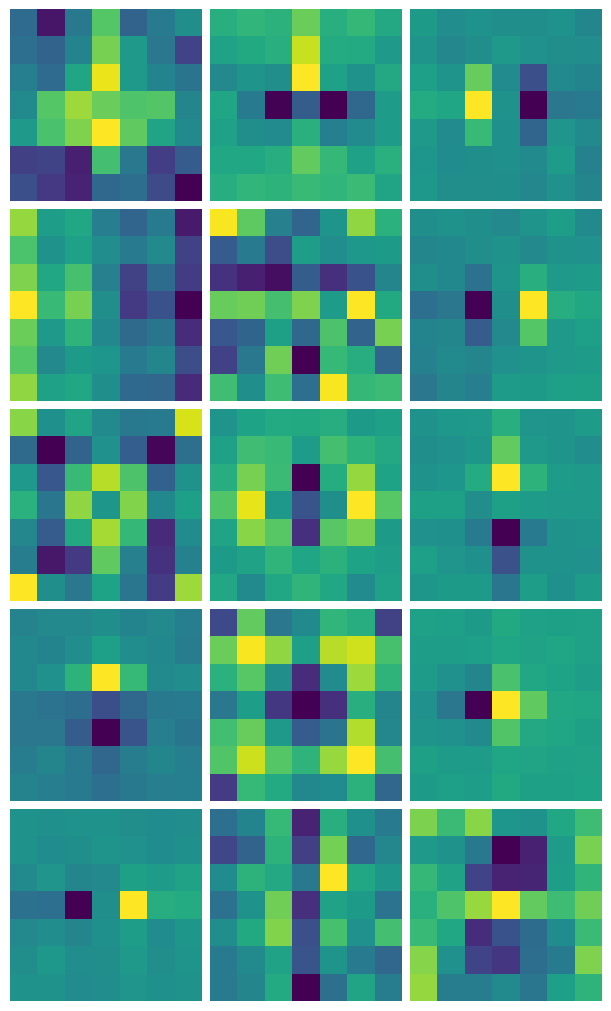}
    \includegraphics[width=0.3227\linewidth]{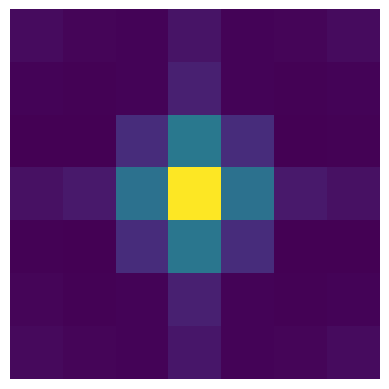}
    \includegraphics[width=0.3227\linewidth]{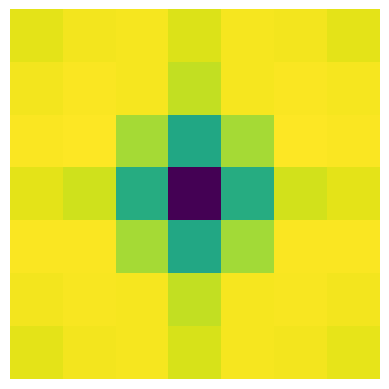}
    \includegraphics[width=0.3227\linewidth]{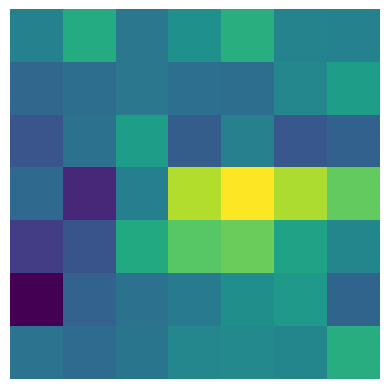}
    \caption{Hornet small, kernel 7×7}
  \end{subfigure}
  \hfill
  \begin{subfigure}{0.24\linewidth}
    \includegraphics[width=0.3227\linewidth]{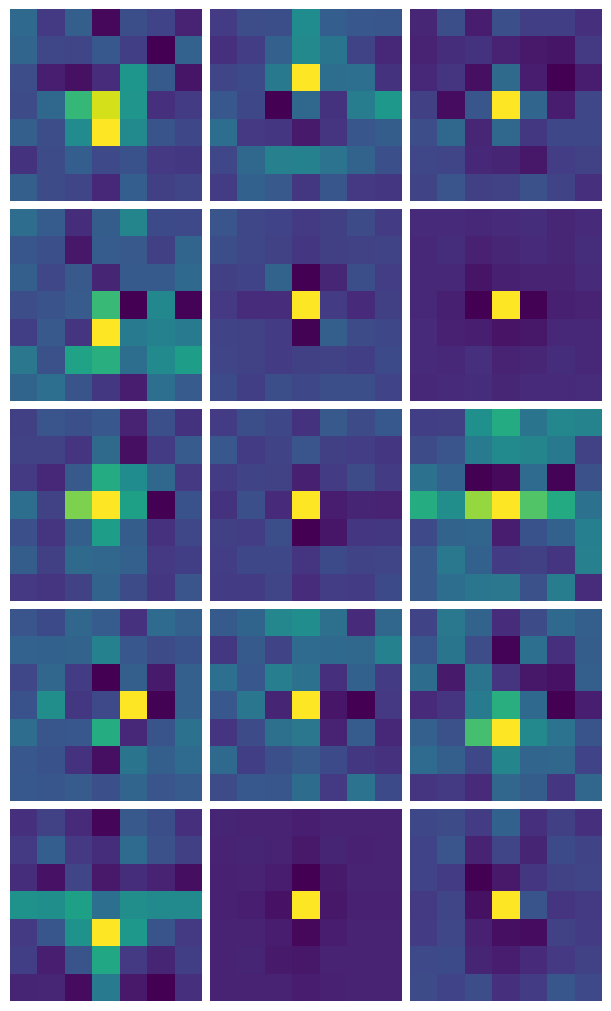}
    \includegraphics[width=0.3227\linewidth]{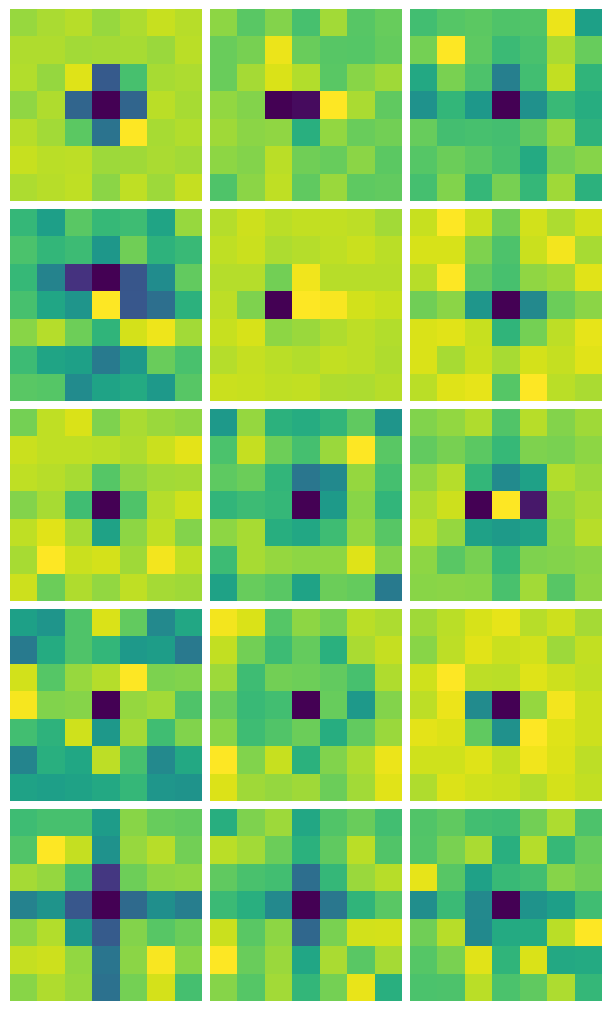}
    \includegraphics[width=0.3227\linewidth]{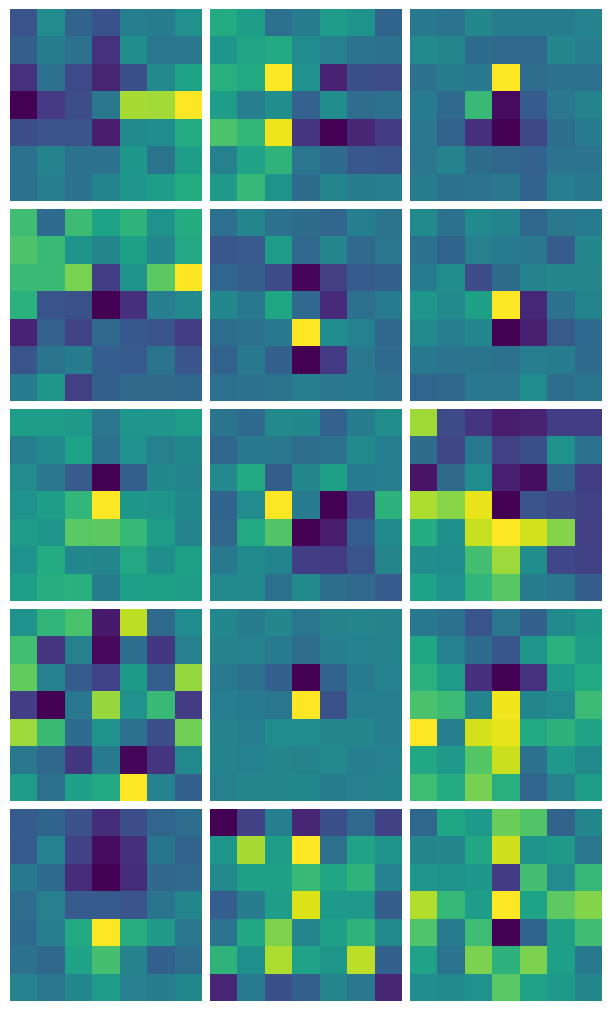}
    \includegraphics[width=0.3227\linewidth]{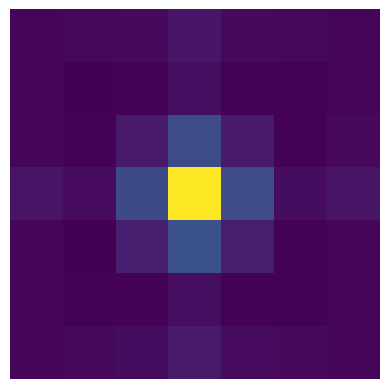}
    \includegraphics[width=0.3227\linewidth]{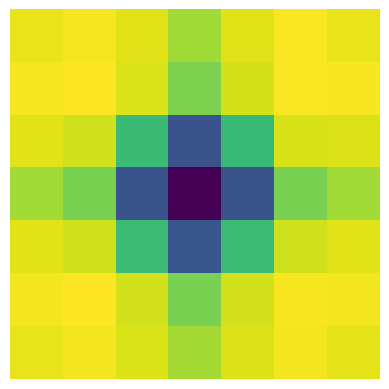}
    \includegraphics[width=0.3227\linewidth]{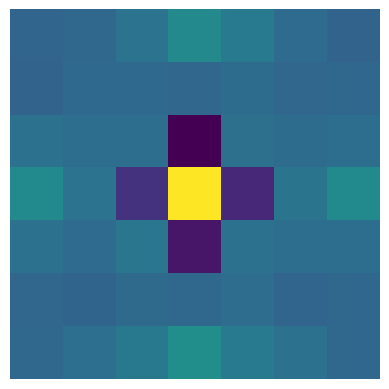}
    \caption{ConvMixer-512, kernel 7×7}
  \end{subfigure}
  \caption{Kernels randomly selected from each K-Means cluster (top) and their respective cluster averages (bottom). Right clusters resemble excitatory-centered fields and middle clusters resemble inhibitory-centered. Clusters on the left contain all other patterns, resulting in their average being cluttered, implying the dominance in the first two cluster patterns.}
   \label{fig:clustered_kernels}
   \vspace{-2ex}
\end{figure*}

\vspace*{0.5ex}\textbf{Initialization Methods.} Kernel initialization methods are crucial in training deep convolutional neural networks (CNNs) and have been the focus of significant research. The initialization of convolutional kernels directly impacts the convergence speed and the final performance of CNNs. Traditional initialization methods include Glorot and Bengio's uniform initialization~\cite{pmlr-v9-glorot10a}, He et al.'s Kaiming initialization~\cite{he2015delving}, and LeCun’s Normal initialization. Mishkin and Matas introduced the LSUV initialization, optimized for deep architectures \cite{init1}. Hanin and Rolnick identified and addressed initialization failure modes in deep ReLU networks \cite{init2}. Arpit et al. proposed a robust initialization for weight normalized and ResNets, demonstrating enhanced generalization in deeper structures \cite{init3}. These methods usually generate weights from a Gaussian or uniform distribution with zero mean and a certain standard deviation. These initialization methods aim to maintain a reasonable activation variance across layers to avoid the issue of vanishing or exploding gradients.

Kaiming initialization, also known as He initialization~\cite{he2015delving}, is a widely adopted technique for initializing weights in convolutional neural networks. Kaiming initialization addresses the vanishing/exploding gradients problem by initializing weights in a way that the variance of the outputs of each convolutional layer is approximately the same as the variance of its inputs. Specifically, the weights are initialized from a Gaussian distribution with a mean of zero and a standard deviation of $\sqrt{2/n}$, where $n$ is the number of inputs to the neuron. This technique has been shown to significantly improve the speed of convergence in deep neural networks and to stabilize the training process.

\section{Methods}

In the following section, we delve into the specifics of our proposed methodology. We begin by conducting a comprehensive analysis of the trained kernels of various state-of-the-art (SOTA) models on the ImageNet dataset. We provide detailed visual illustrations coupled with quantitative results, to validate the presence of the center-surround antagonism in a considerable portion of the kernels.

Next, we move on to discuss the Difference of Gaussian (DoG) function. This function serves as a mathematical model of the center-surround receptive fields found in biological visual systems. This mathematical model provides us with a foundation for designing an initialization scheme that mimics these biological structures.

Finally, we describe our novel kernel initialization approach, which leverages the DoG function. We detail the process of applying this function to generate kernel weights that resemble the center-surround antagonism of biological vision systems. The ultimate goal is to provide the model with a starting point that is already attuned to the kind of spatial feature mappings it would otherwise have to learn through many epochs of training.

\subsection{Inspecting the Trained Kernels}

In this section, we conduct a detailed exploration of the trained kernels of the regular and depthwise convolutions in different models. Specifically, we examine VGG16~\cite{DBLP:journals/corr/SimonyanZ14a}, ResNet50~\cite{he2016residual}, DenseNet201~\cite{DenseNet}, MobilenetV3~\cite{mobilenetv3}, EfficientNet~\cite{efficientnet}, ConvNeXt~\cite{convnext}, ConvNeXtV2~\cite{ConvNeXtV2}, Hornet~\cite{hornet}, and ConvMixer~\cite{convmixer}. For our analysis, we utilized the pre-trained versions of these models, sourced directly from Pytorch or their respective official code repositories.

\begin{figure}[h]
  \centering
  \begin{subfigure}{0.31\linewidth}
    \includegraphics[width=\linewidth]{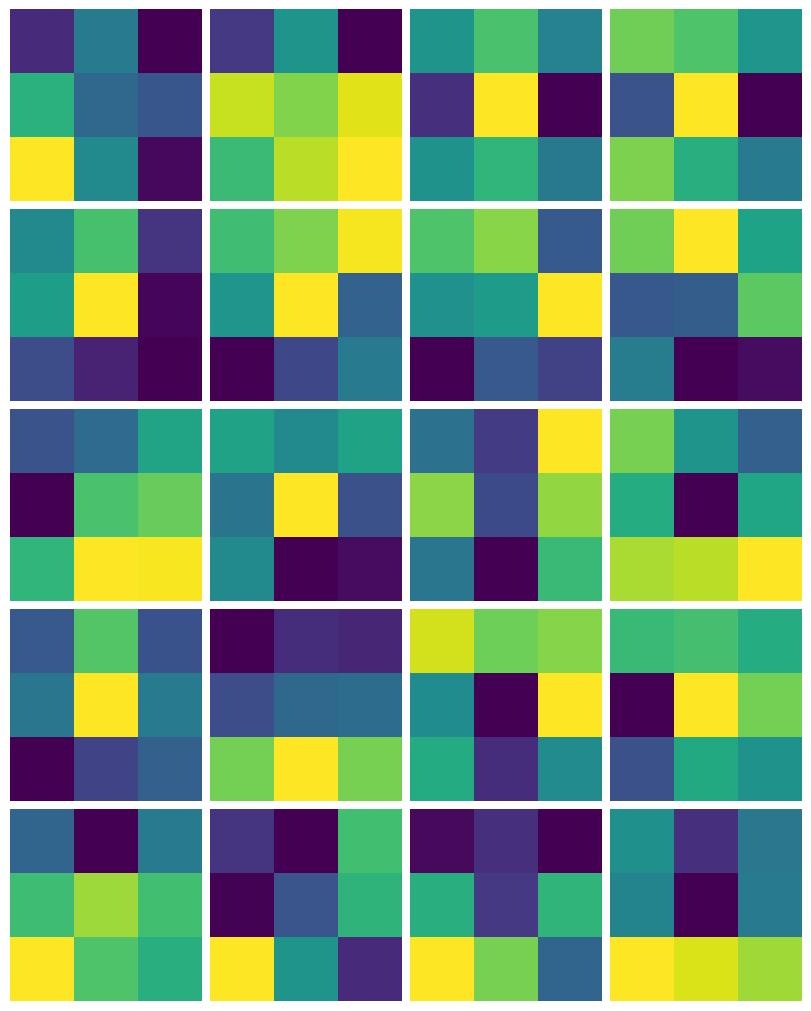}
    \caption{ResNet50}
  \end{subfigure}
  \hfill
  \begin{subfigure}{0.31\linewidth}
    \includegraphics[width=\linewidth]{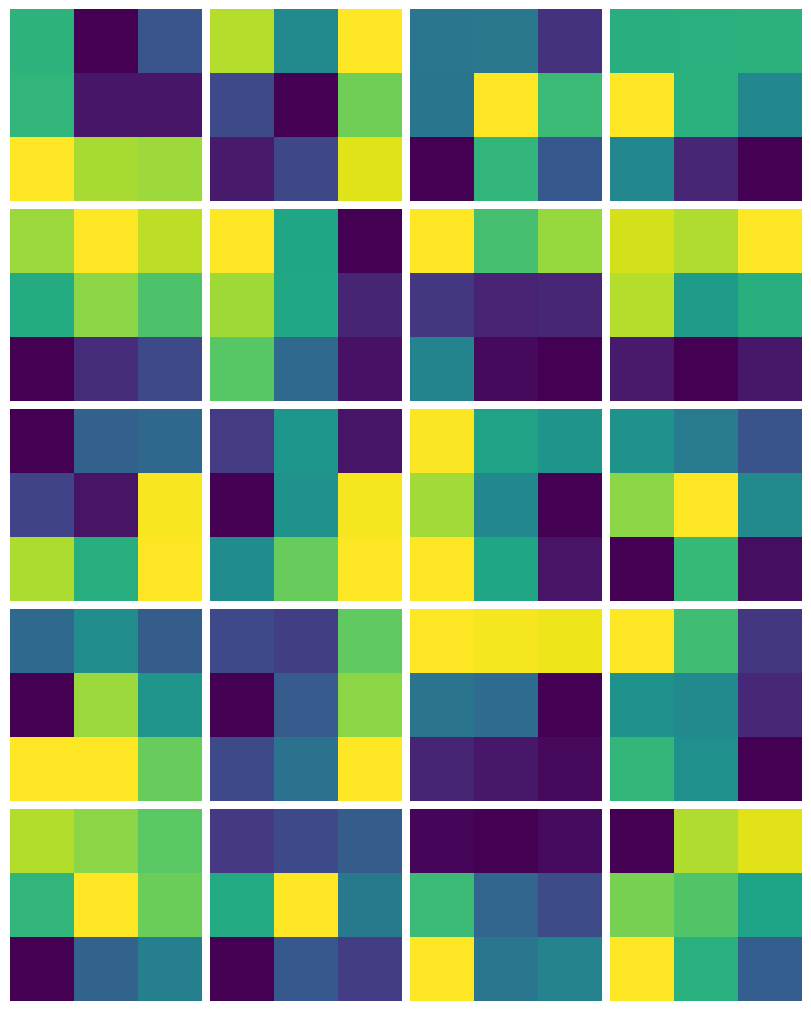}
    \caption{VGG16}
    \end{subfigure}
    \hfill
  \begin{subfigure}{0.31\linewidth}
    \includegraphics[width=\linewidth]{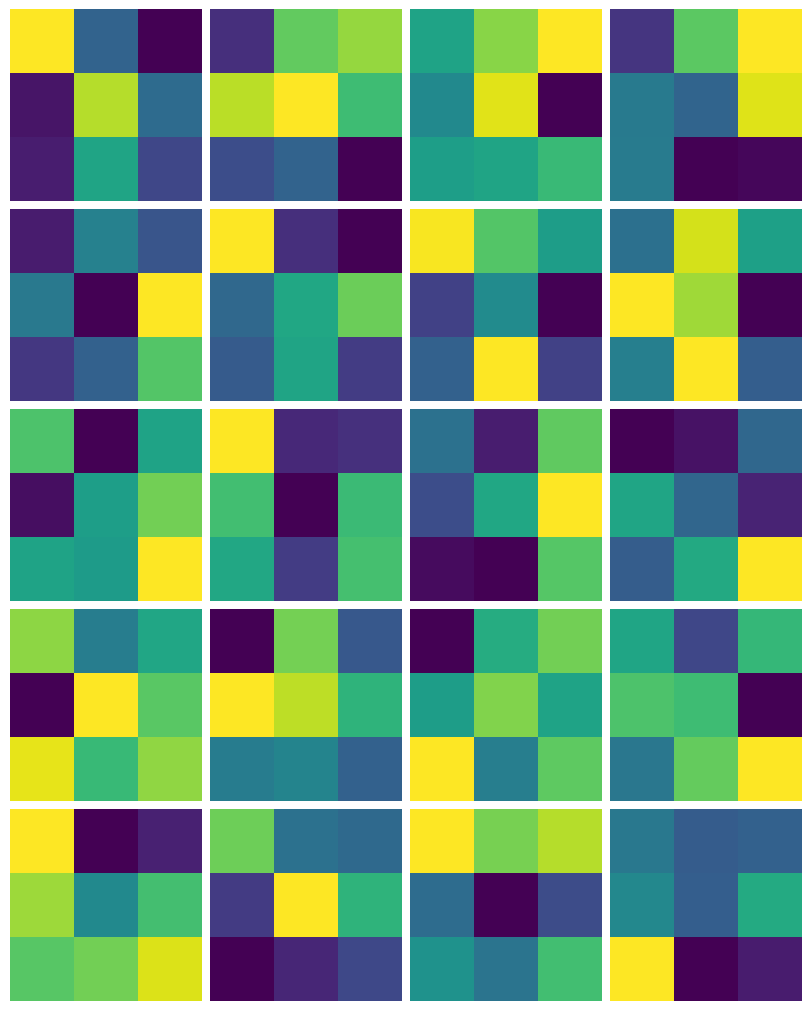}
    \caption{DenseNet201}
  \end{subfigure}

  \vspace{-0.5ex}
  \caption{Random samples from regular convolutions of popular models, trained on the ImageNet dataset. Unlike depthwise convolutions, kernels from regular convolutions do not have visually observable repeated patterns.}
  \label{fig:Mixed_kernels_reg}
  \vspace{-2ex}
\end{figure}

In order to visually inspect the learned patterns within these filters, we have included Figures~\ref{fig:Mixed_kernels} and ~\ref{fig:Mixed_kernels_reg} in our paper. Figure~\ref{fig:Mixed_kernels} provides a representative selection of randomly chosen samples from the trained kernels of each of the models utilizing depth-wise convolutions, and Figure~\ref{fig:Mixed_kernels_reg} shows the same for models with regular convolutions. Upon inspection of these kernels, we can identify recurring patterns among the depthwise convolutions. We observed similar patterns in other variants of these models trained on ImageNet, too. However, kernels of regular convolutions do not show such observable patterns

One particularly notable pattern in depthwise kernels is the center-focused structure of many of them. Interestingly, these center-focused kernels can be broadly divided into two categories. The first category includes kernels with larger weight values concentrated in their center. Conversely, the second category consists of kernels with larger weight values populating their surround.

These discovered patterns bear striking resemblance to the well-studied `center-surround antagonism' found in the mammalian visual system which we discussed previously.

Nevertheless, it's worth noting that not all filters from our models exhibit this center-surround pattern. We observed other filters possessing different, non-center-surround patterns, but their occurrence was less frequent compared to the center-surround ones. This differential frequency points to the significance and prevalence of the center-surround pattern in the learned representations of depthwise kernels in these models.

\begin{figure}
  \centering
  \begin{subfigure}{0.325\linewidth}
    \includegraphics[width=0.316\linewidth]{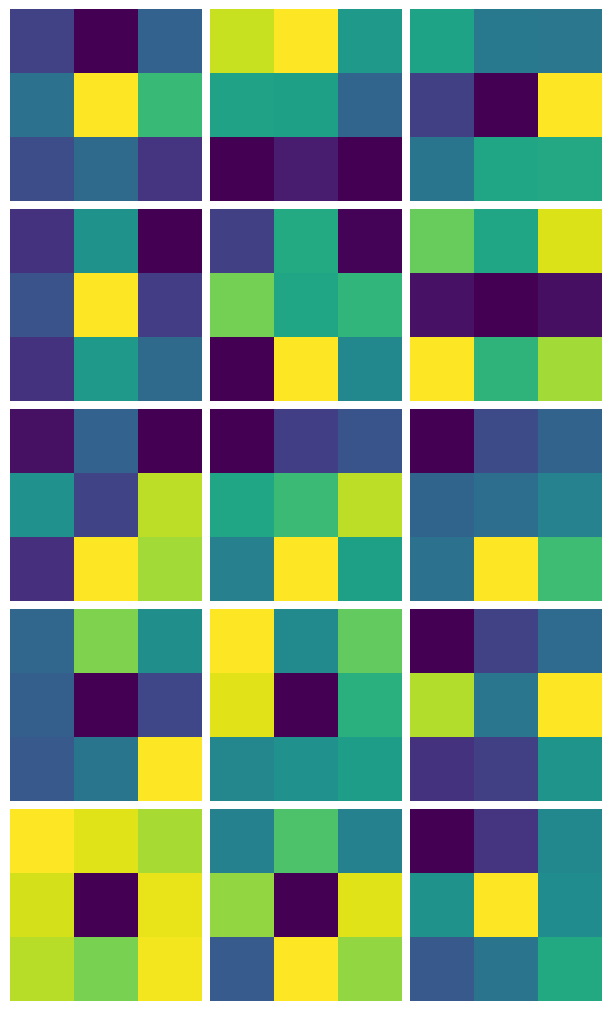}
    \includegraphics[width=0.316\linewidth]{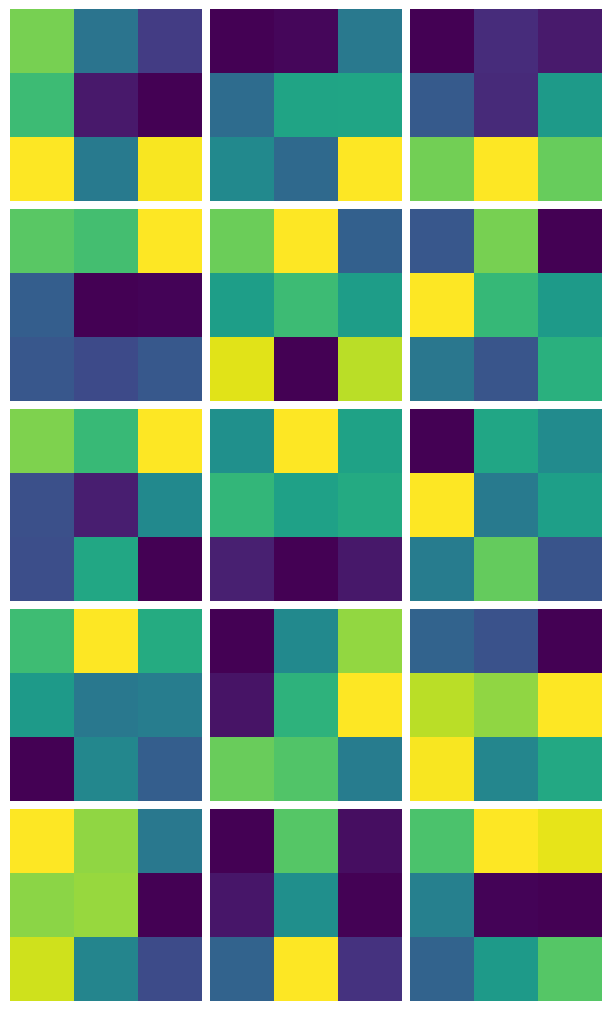}
    \includegraphics[width=0.316\linewidth]{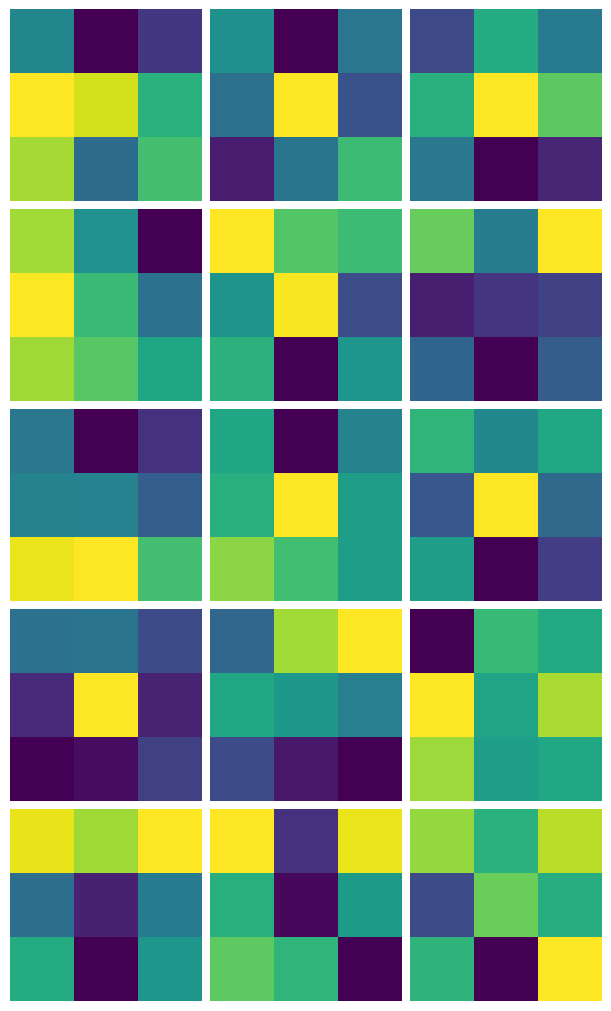}
    \includegraphics[width=0.316\linewidth]{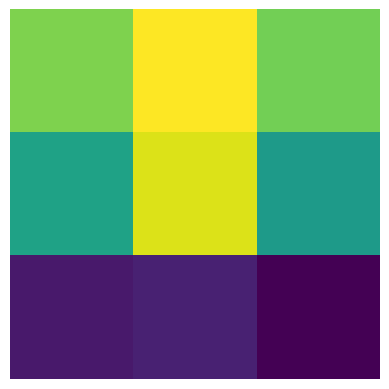}
    \includegraphics[width=0.316\linewidth]{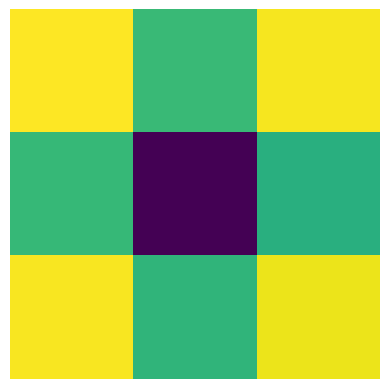}
    \includegraphics[width=0.316\linewidth]{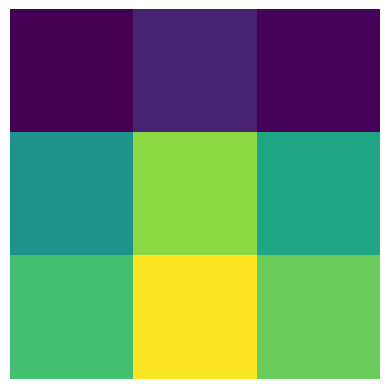}
    \caption{ResNet50}
  \end{subfigure}
  \hfill
  \begin{subfigure}{0.325\linewidth}
    \includegraphics[width=0.316\linewidth]{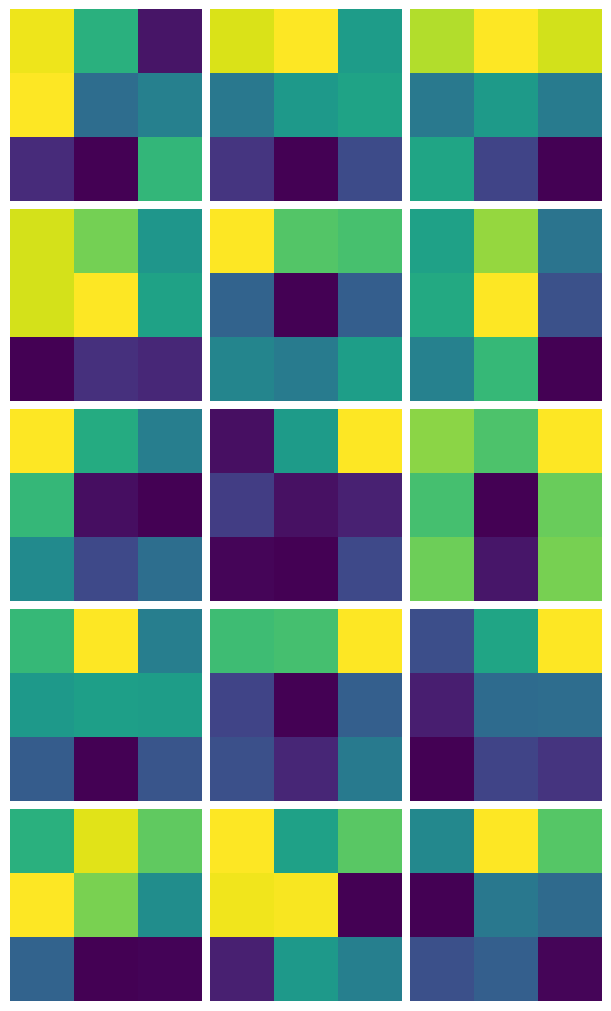}
    \includegraphics[width=0.316\linewidth]{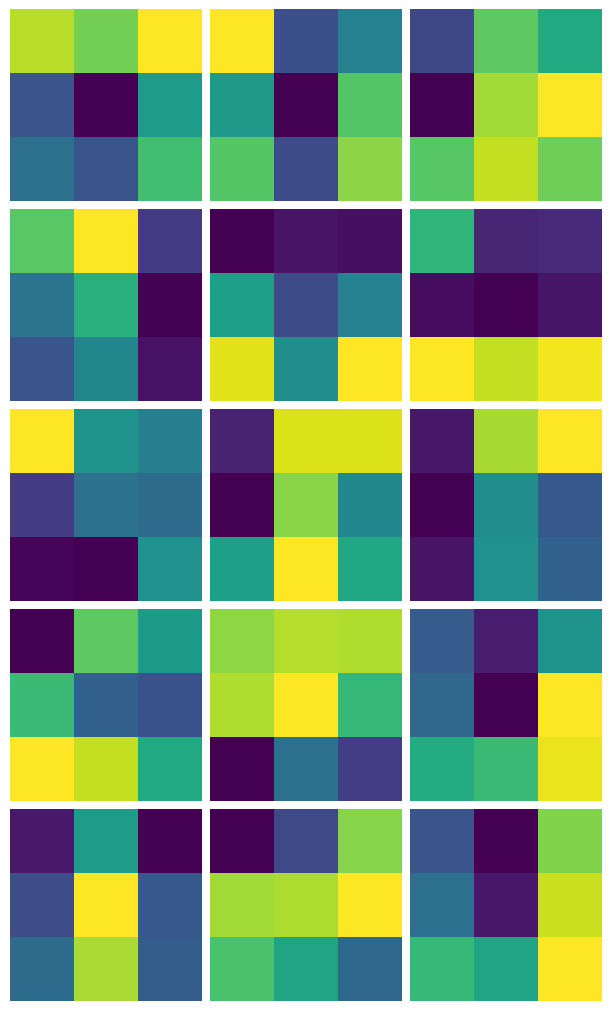}
    \includegraphics[width=0.316\linewidth]{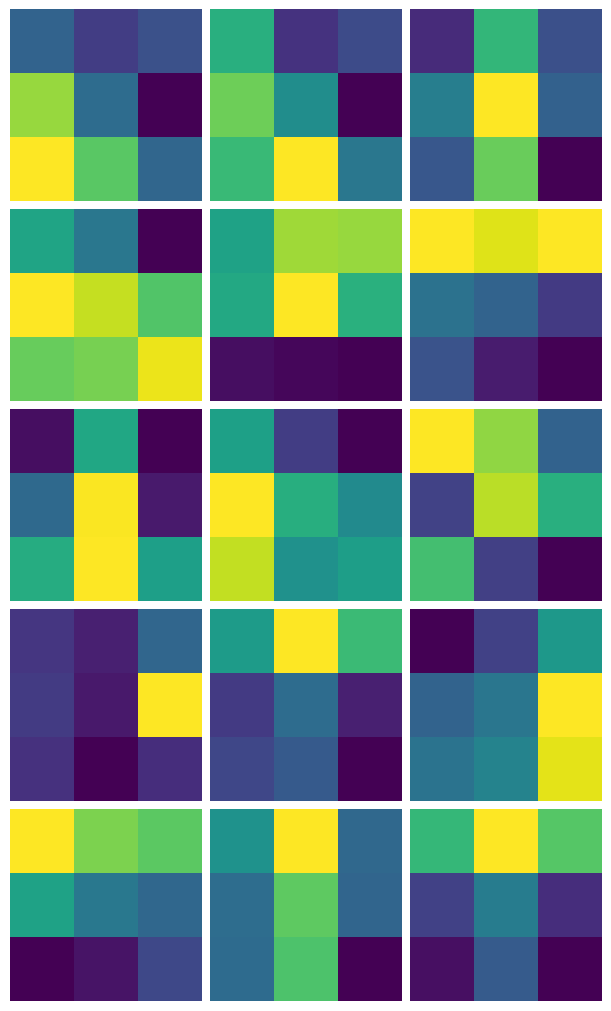}
    \includegraphics[width=0.316\linewidth]{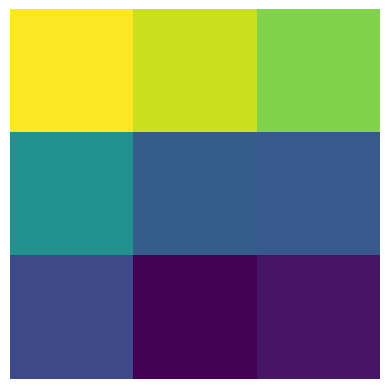}
    \includegraphics[width=0.316\linewidth]{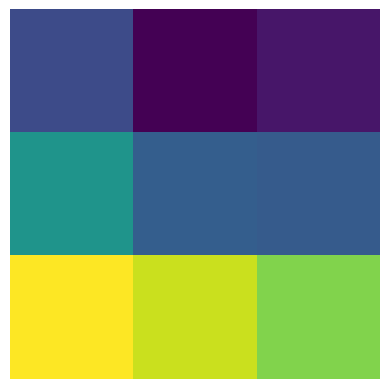}
    \includegraphics[width=0.316\linewidth]{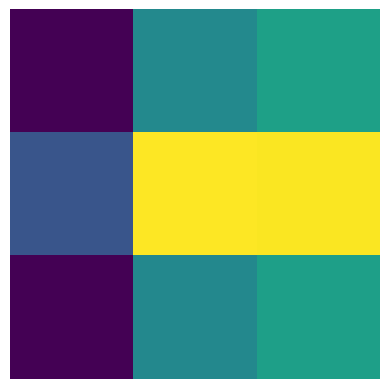}
    \caption{VGG16}
  \end{subfigure}
  \hfill
  \begin{subfigure}{0.325\linewidth}
    \includegraphics[width=0.316\linewidth]{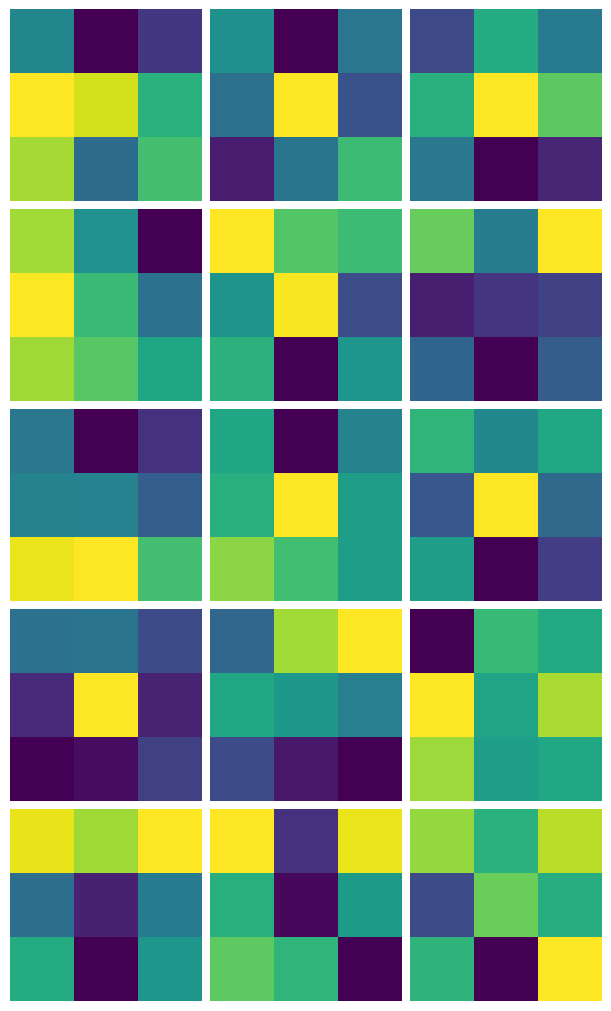}
    \includegraphics[width=0.316\linewidth]{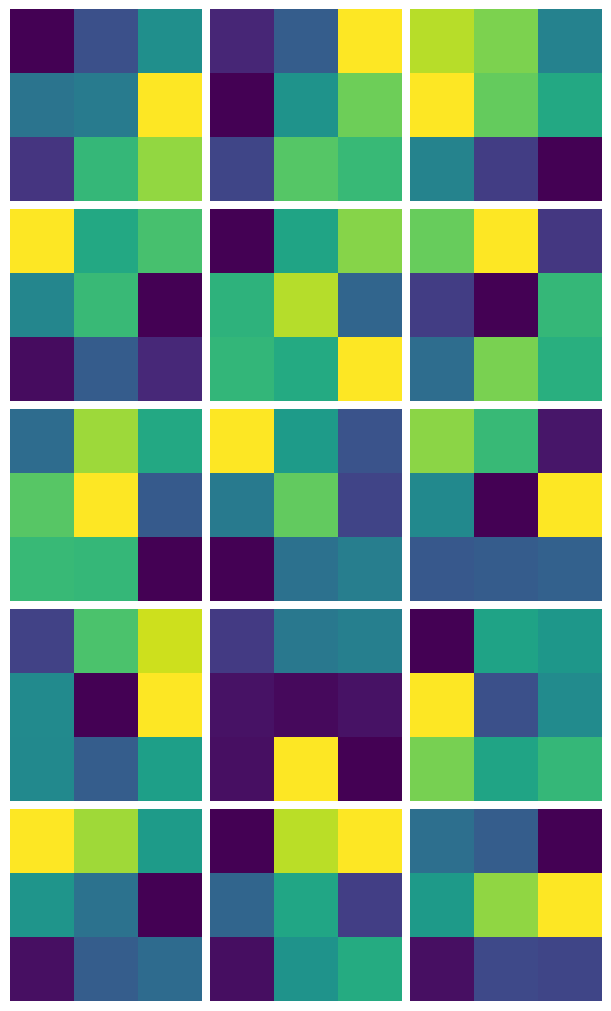}
    \includegraphics[width=0.316\linewidth]{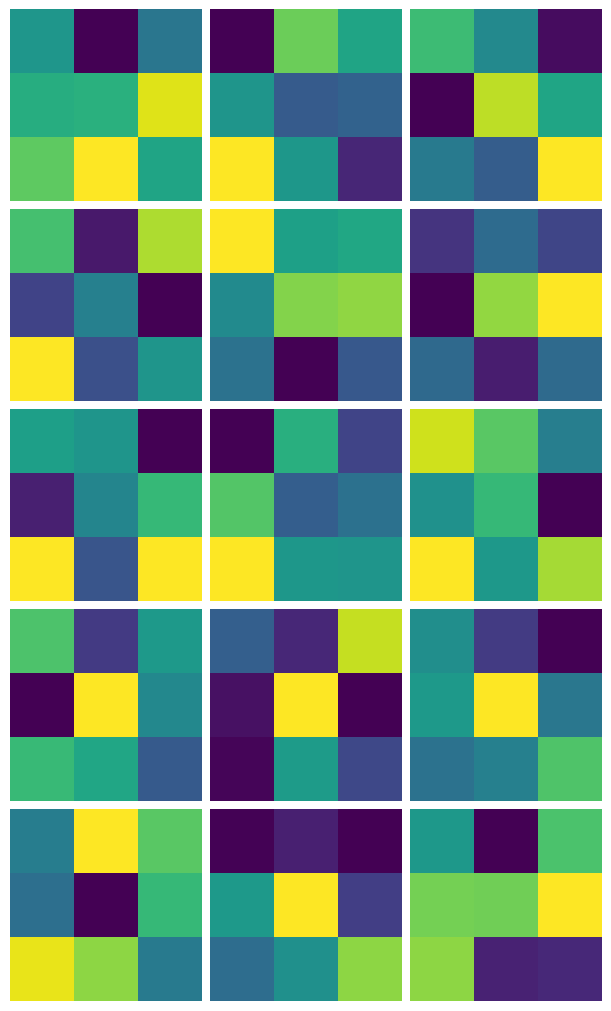}
    \includegraphics[width=0.316\linewidth]{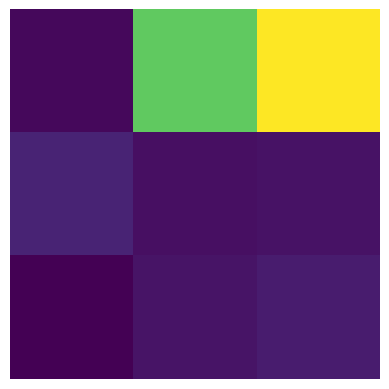}
    \includegraphics[width=0.316\linewidth]{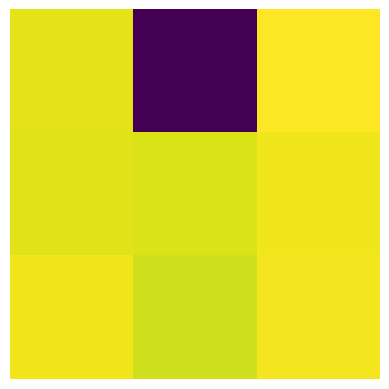}
    \includegraphics[width=0.316\linewidth]{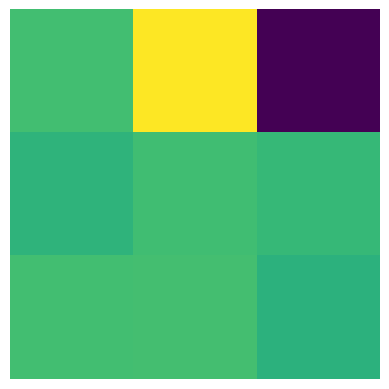}
    \caption{DenseNet201}
  \end{subfigure}
  
  \caption{Clusters of 3×3 kernels of regular convolutions}
  \label{fig:clusters_reg}
  \vspace{-2ex}
\end{figure}

To attain a more analytic comprehension of the varying kernel patterns, we leverage a straightforward clustering algorithm to group all the kernels. Our hypothesis suggests that the two center-surround groups are the most significant patterns, thus we set the number of clusters to three in the clustering algorithm. This choice is to discern whether the algorithm can effectively categorize the kernels into two distinct center-surround clusters, and a third cluster comprising the less common patterns.

For a successful execution of clustering, we took some preparatory steps. First, we normalized all kernels to have their weight values lie within the range of 0 and 1, utilizing min-max encoding. This step is essential to ensure the numerical stability and effectiveness of the clustering algorithm. Following normalization, we flattened each kernel into a vector, to fit the input requirement of the k-means algorithm. With the transformed data, we were finally able to run the k-means algorithm~\cite{Hartigan1979} with a k-value of 3.

After clustering, we visually inspected the kernels belonging to each cluster, by presenting randomly selected samples in Figure~\ref{fig:clustered_kernels}. For each cluster, we also depict the average of all kernels belonging to it. This helps in better seeing the prominent pattern of each cluster. The first cluster predominantly contained kernels akin to the excitatory-center receptive fields, while the second cluster closely resembled inhibitory-center receptive fields. As for the third cluster, the kernels exhibited some degree of center-focused structures but lacked the precise characteristics of center-surround receptive fields. Examination of the average kernel within each cluster reveals a pronounced center-surround structure in the first two clusters, whereas the third cluster exhibits a more dispersed pattern. This observation further underscores that even an unbiased clustering approach distinctly recognizes the prominence of center-surround patterns relative to alternative patterns.

\begin{figure}[h]
  \centering
  \begin{subfigure}{0.37\linewidth}
    \includegraphics[width=0.316\linewidth]{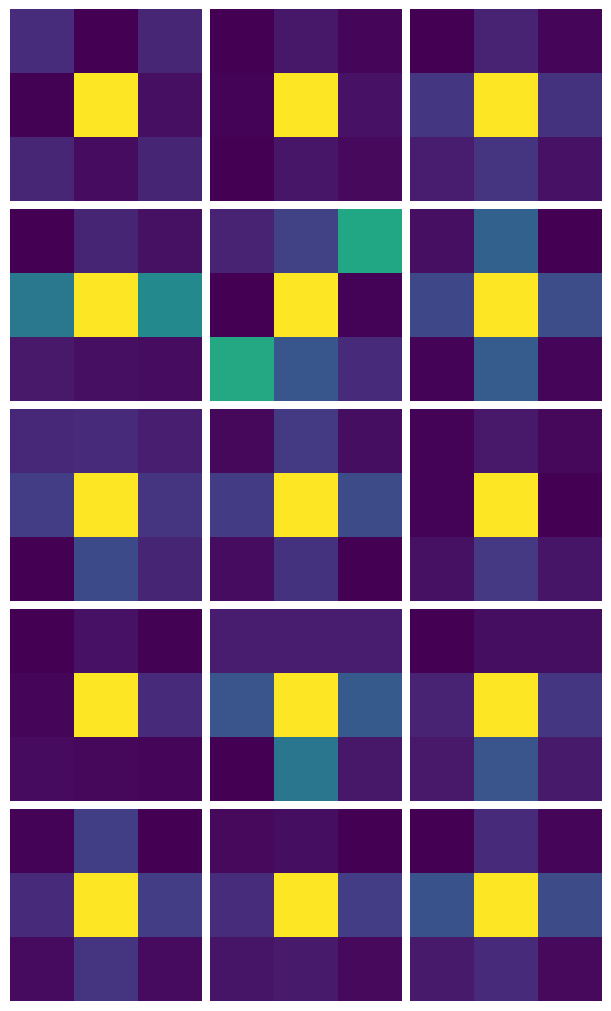}
    \includegraphics[width=0.316\linewidth]{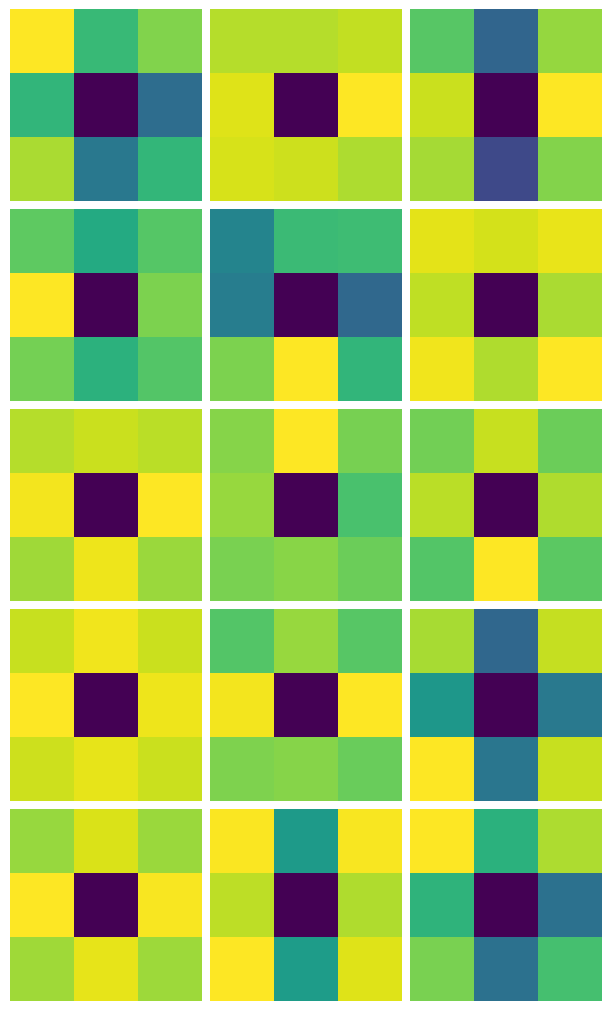}
    \includegraphics[width=0.316\linewidth]{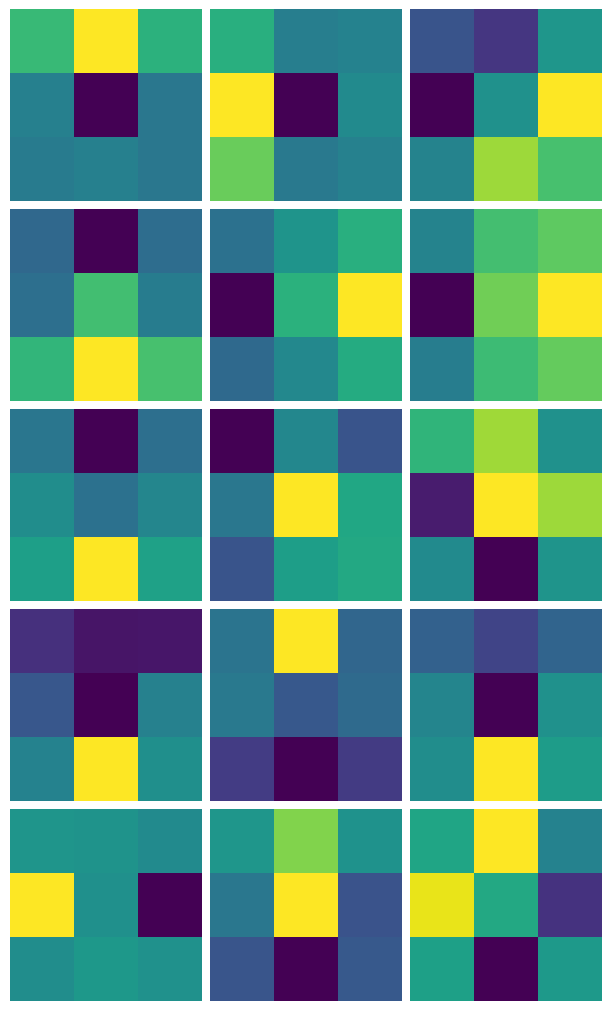}
    \includegraphics[width=0.316\linewidth]{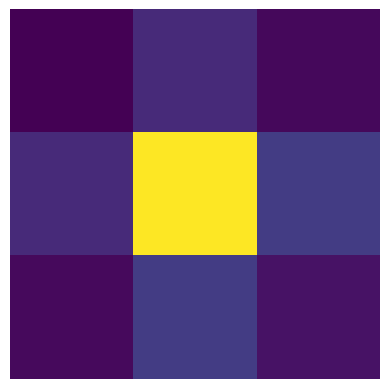}
    \includegraphics[width=0.316\linewidth]{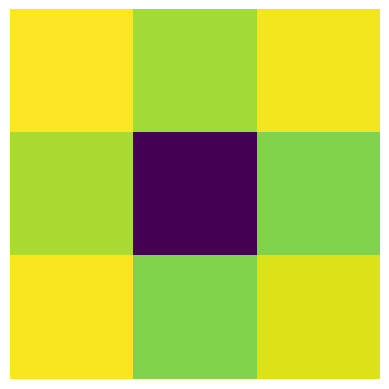}
    \includegraphics[width=0.316\linewidth]{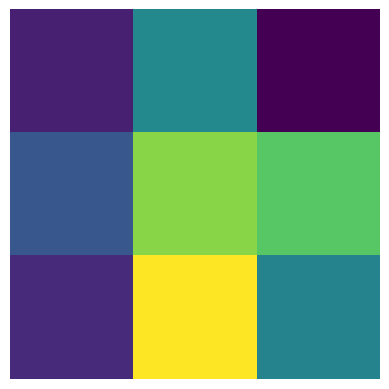}
    \caption{MobileNetV3}
  \end{subfigure}
  \hspace{0.7cm}
  \begin{subfigure}{0.37\linewidth}
    \includegraphics[width=0.316\linewidth]{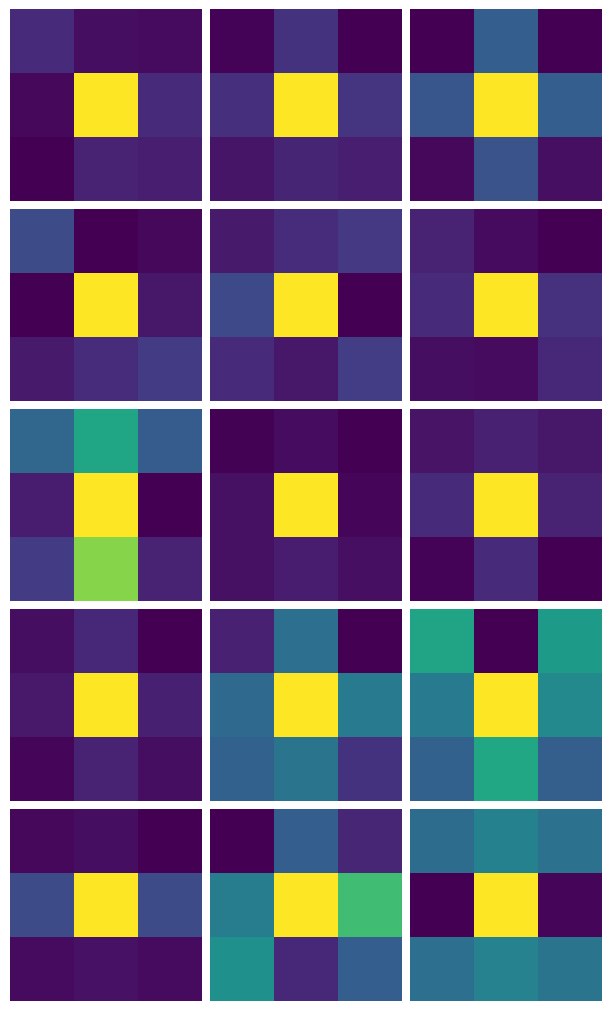}
    \includegraphics[width=0.316\linewidth]{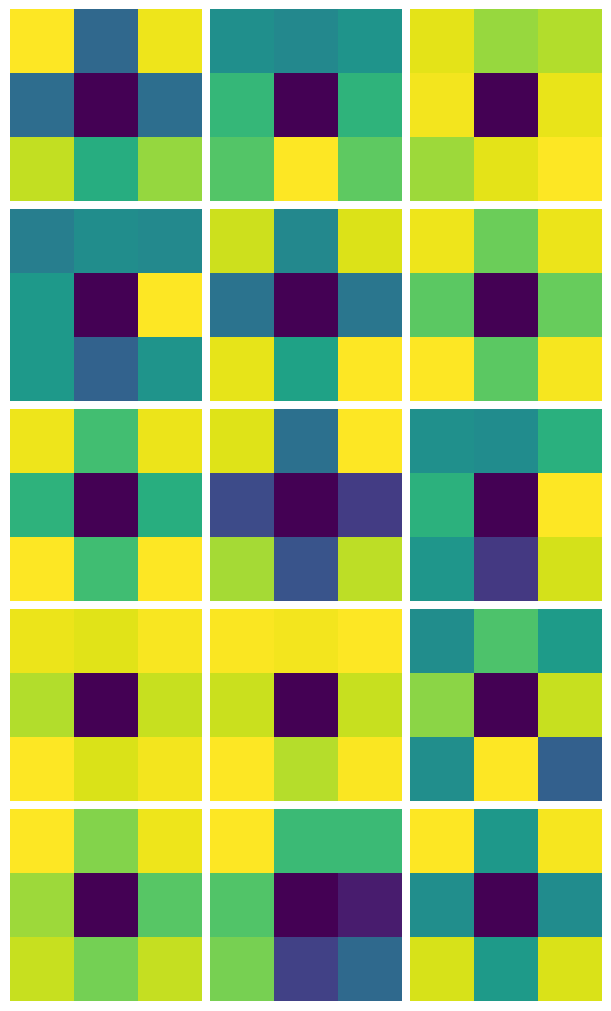}
    \includegraphics[width=0.316\linewidth]{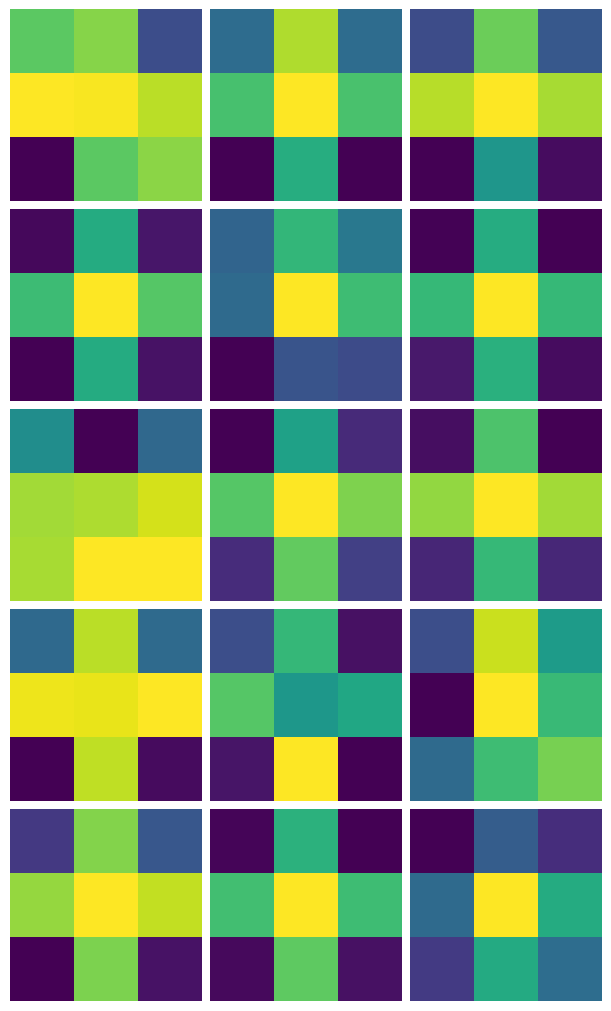}
    \includegraphics[width=0.316\linewidth]{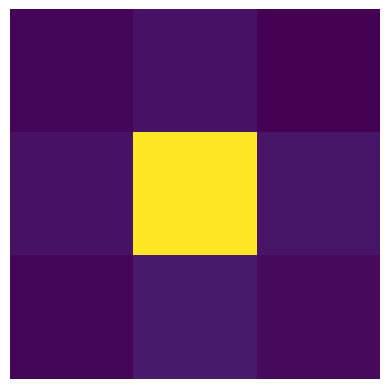}
    \includegraphics[width=0.316\linewidth]{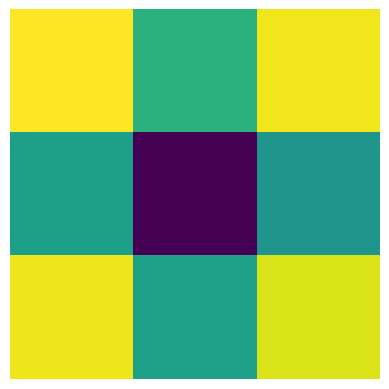}
    \includegraphics[width=0.316\linewidth]{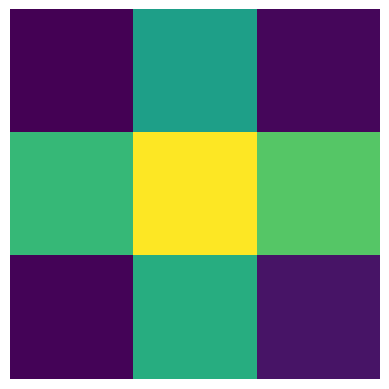}
    \caption{EfficientNet}
  \end{subfigure}
  
  \caption{Clusters of 3×3 kernels of depthwise convolutions}
  \label{fig:clusters_dp}
  \vspace{-2ex}
\end{figure}

Figures~\ref{fig:clusters_reg} and ~\ref{fig:clusters_dp} show the discovered clusters of the models with $3\times3$ kernels with regular and depthwise convolutions, respectively. As one can see, the models with small depthwise kernels still have prominent center-surround clusters, in contrast to the ones with regular convolutions. Only in Resnet50, the average kernels of one cluster is similar to the center-surround pattern. However, once inspecting the kernels themselves, we can not distinguish the patterns (See Figure~\ref{fig:Mixed_kernels_reg}). 

\begin{figure*}
  \centering
   \includegraphics[width=0.98\linewidth]{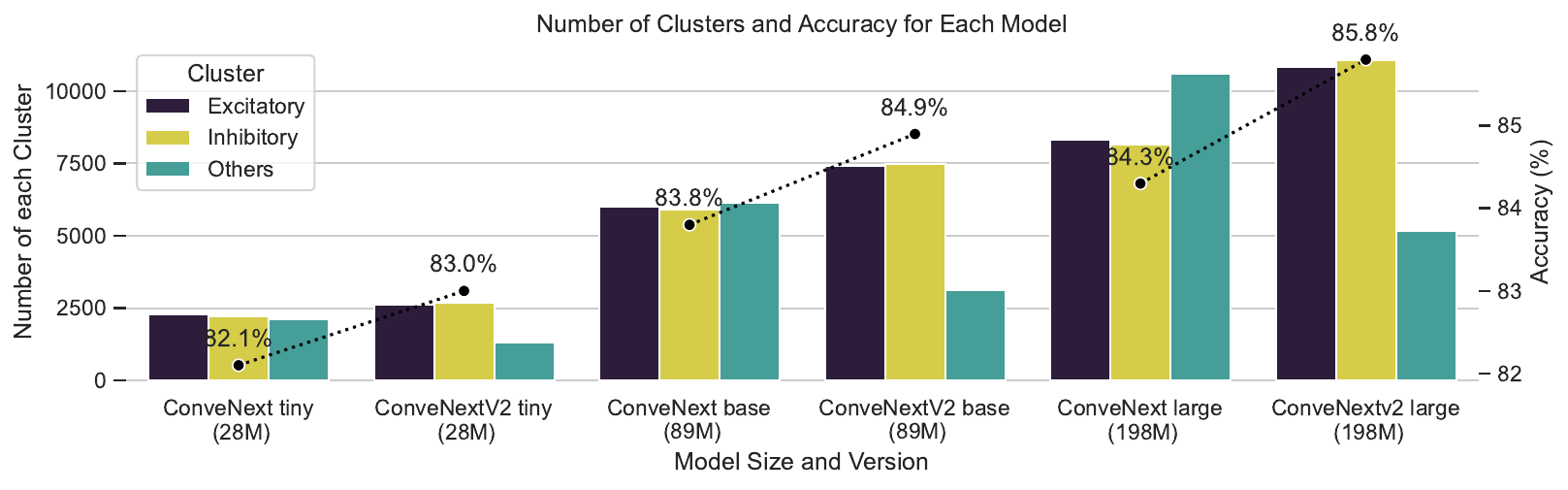}
   \caption{Histograms of the clusters discovered in ConvNeXt model variants, alongside their V2 counterpart, including the test accuracy of each model. The improved V2 versions have a considerably lower number of kernels in their "others" cluster.}
   \label{fig:histograms}
   \vspace{-2.5ex}
\end{figure*}

The distribution of the kernels across each cluster of ConvNeXt compared to ConvNeXtV2 variants is demonstrated by the histograms shown in Figure~\ref{fig:histograms}. The ConvNeXt model faced challenges with feature collapse, manifesting as redundant activations across channels. In response, the ConvNeXt V2 architecture introduces the Global Response Normalization (GRN) layer, promoting feature diversity. This enhancement, coupled with advanced self-supervised techniques, positions ConvNeXtV2 as a marked improvement over its predecessor in visual recognition tasks. We note a significant reduction in the number of kernels within the third cluster of the improved ConvNeXtV2 when contrasted with its predecessor, ConvNeXtV1. This is an interesting observation that underscores the pivotal role of center-surround in enhancing the model's performance.

\begin{figure*}[t]
  \centering
   \includegraphics[width=\linewidth]{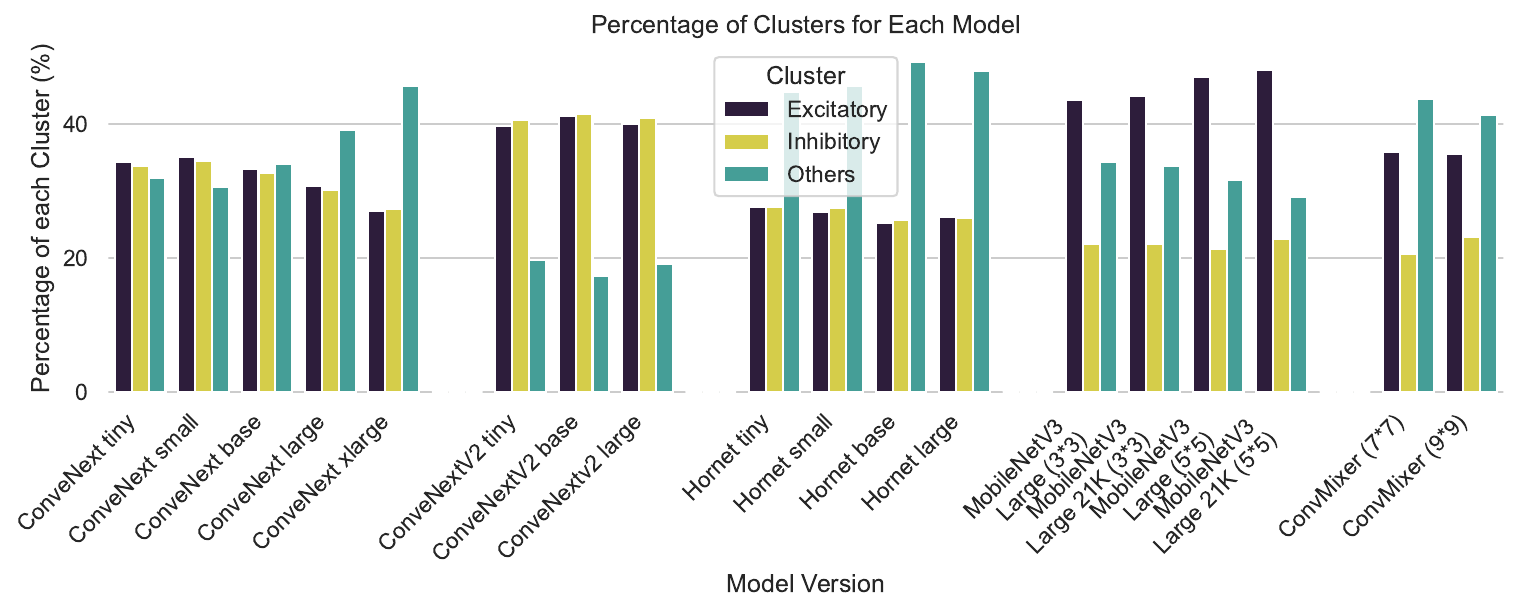}
   \caption{Filter Proportions by Cluster in Various Models: We observe almost consistent proportions of filter clusters across models of different sizes, kernel sizes, and dataset sizes within each model category.}
   \label{fig:histograms_2}
   \vspace{-2.5ex}
\end{figure*}

As detailed in Figure~\ref{fig:histograms_2}, we observed a remarkable consistency in the proportions of filter clusters across various models, despite changes in model sizes, kernel sizes, and dataset sizes. This trend was evident in models such as ConvNeXt, ConvNeXtV2, and Hornet, where different model sizes were analyzed. For MobileNet and ConvMixer, we extended this analysis to include variations in kernel sizes. Additionally, the training of MobileNet on both ImageNet 1K and 21K datasets did not significantly alter the proportion of filter clusters. These findings suggest an inherent stability in the distribution of filter types within each model category, indicating that the architectural design of these models plays a more critical role in determining filter distribution than the scale of the model or the size of the dataset. This insight could have implications for understanding the scalability and adaptability of these models to different sizes and types of datasets.


\subsection{Formulation of Center-Surround Kernels} 
The computation of center and surround weights can be accomplished using a difference of two Gaussian functions (DoG). Represented in Cartesian coordinates (CC), with the CC origin designated as the receptive field's center, the DoG can be formulated as presented in Rodieck's work \cite{rodieck_1965}:
\begin{equation}
\centering
\label{eq:rodiek}
DoG(x,y) = K_1 e^{-\frac{x^2+y^2}{\sigma_1}} - K_2
e^{-\frac{x^2+y^2}{\sigma_2}}
\end{equation}
where it holds true that $K_1,{>},K_2$ and $\sigma_2,{>},\sigma_1$~\cite{Blackburn1993ASC}.

We use the DoG model proposed by Petkov and Kruizinga \cite{Petkov2005ModificationsOC,kruizinga_petkov_2000}, which defines the difference of Gaussians for the center and surround kernels. This model enables us to calculate the variances analytically, given the kernel size and the ratio of the center to surround \cite{Petkov2005ModificationsOC}:
\begin{equation}
\label{eq:petrov}
DoG_{\sigma,\gamma}(x,y) = \frac{A_c}{\gamma^2} e^{-\frac{x^2+y^2}{2\gamma^2\sigma^2}} - A_s e^{-\frac{x^2+y^2}{2\sigma^2}}
\end{equation}

In this formula, $\gamma\,{<}\,1$ stipulates the ratio between the center radius $r$ and the surround. The coefficients $A_c$ and $A_s$ are determined by requiring the sum of all positive values in Equation~\ref{eq:petrov} to be equivalent to the negative values. These are then normalized such that their sum equals 0.5 and -0.5, respectively. While in the continuous infinite case, the coefficients $A_c$ and $A_s$ are equal, in the discrete finite case, the values of $A_c$ and $A_s$ remain remarkably similar.

By setting $DoG_{\sigma,\gamma}(x,y) = 0$, $\sigma$ can be calculated immediately as shown in Equation~\ref{eq:sigma} below, where $k$ is the kernel size, for any arbitrary values of $k$ and $\gamma$:
\begin{equation}
\label{eq:sigma}
\sigma \approx \frac{k}{4}\sqrt{\frac{1-\gamma^2}{-\ln{\gamma}}}
\end{equation}
In Figure~\ref{fig:DoG}, we show the DoG functions used to model the center-surround receptive fields, with either excitatory or inhibitory centers, respectively. 

\subsection{Center-Surround Initialization}

Observing the repetitive patterns exhibited in the depthwise kernels trained on ImageNet, and noting their resemblance to center-surround receptive fields, we propose a novel methodology for kernel-weight initialization. Our hypothesis is based on the assumption that by offering the model kernels an initialization that aligns with patterns not only found in nature but also in fully-trained models, we can enhance the performance of these models. Additionally, this approach might streamline the convergence process during training. This method potentially serves as a bridge, linking biological vision models with their artificial counterparts, thereby enabling the latter to benefit from the intrinsic efficiency of the former. 

In our approach, we begin by initializing the weights of the depth-wise convolution kernels in the model architectures with the weights derived from the previously discussed DoG function. This is a crucial step that enables us to effectively incorporate the center-surround receptive field structure into the model. To achieve a balanced representation of both inhibitory and excitatory centers, each kernel is assigned a center type – either inhibitory or excitatory – with an equal probability of 50\%. This ensures that both types of centers are represented in approximately equal proportions across the kernels, thus maintaining a balanced interaction of these opposing neural behaviors in the model.

Additionally, we introduce variability in the ratio of the center to the surround for each kernel. To do this, we select the ratio from a uniform distribution. This introduces an element of randomness to the model, ensuring that a wide variety of center-to-surround ratios are represented in the kernels. Consequently, this design enables the model to accommodate and respond to a broad range of spatial scales in the input data and adds a wider variety of weight values to the initialization.

Figure~\ref{fig:kernel_ratios} shows the DoG kernels of size 9 with different ratios of the center to surround varying from 0.2 to 0.8. As the ratio increases, the center gets larger.

\section{Experiments}
Here, we detail our implementation, covering model selection and training. We then showcase results from testing our initialization on ConvNeXt, HorNet, and ConvMixer models using Cifar10~\cite{cifar10} and ImageNet~\cite{DBLP:conf/cvpr/DengDSLL009}. Lastly, we provide an ablation study on our approach's facets.

\subsection{Implementation Details}
For the ImageNet evaluations, we employed ConvNeXt tiny, HorNet Tiny, and ConvMixer-512. ConvNext tiny and HorNet tiny contain 18 and 25 blocks respectively, each composed of one depthwise and two pointwise convolutions. The ConvMixer models incorporate 512 filters, within each depthwise convolutional layer and comprise mixer blocks composed of depthwise and pointwise convolutions. For each model, we used the training settings proposed in the original paper. Our training regimen included the use of a suite of data augmentation techniques, namely RandAugment~\cite{cubuk2019randaugment}, mixup~\cite{zhang2018mixup}, CutMix~\cite{Yun_2019_ICCV}, and random erasing~\cite{Zhong_Zheng_Kang_Li_Yang_2020}, in addition to gradient norm clipping. We employed the Adam optimizer~\cite{DBLP:journals/corr/KingmaB14} for the training process. 


Across all experimental evaluations, we adhered to a balanced strategy for our kernel initialization, assigning half of the kernels with excitatory centers and the other half with inhibitory centers. Moreover, to determine the value of $\gamma$, representing the ratio of center to surround, we utilized a uniform distribution inside [0,0.5]. This choice is motivated by our observations derived from the trained filters, which showed us that the centers are usually quite small.

\begin{table*}
\small
  \centering
  \caption{ Results of Depthwise Convolutional Models on ImageNet with different settings and initializations.}
 
\begin{tabular}{lccccc}
    \toprule

    Model & kernel Size   & Kaiming Initialization & Our Initialization  \\   
    \midrule
    ConvNeXt tiny & $7\times7$   & 76.17  & \textbf{76.74} \\
    HorNet tiny & $7\times7$ & 76.06 & \textbf{76.40} \\
    ConvMixer-512 & $9\times9$  & 64.00 & \textbf{66.34} \\
    \bottomrule
  \end{tabular}
  \label{tb:ImageNet}
\vspace{-3ex}
  
\end{table*}

\subsection{Results}
In the following, we describe our experimental results on ImageNet and Cifar10 datasets. We compare our results to the Kaiming initialization, which is the default initialization method used in most of the models. 

\vspace*{0.5ex}\textbf{ImageNet.}
In Table~\ref{tb:ImageNet} we present the empirical results of our experiments on ImageNet. Across all configurations, our initialization consistently outperforms the Kaiming initialization, with improvements ranging from marginal to substantial. This was particularly evident in the improvement exceeding 2\% for the ConvMixer-512 with kernel-size $9{\times}9$.

First, we present the performance metrics for the ConvNeXt tiny model, utilizing a $7\times7$ kernel size and trained over 50 epochs. Employing the conventional Kaiming initialization, the model achieved an accuracy of 76.17\%. However, when initialized with our proposed method, the accuracy exhibited a slight enhancement, reaching 76.74\%.

The subsequent row provides the results for the HorNet tiny model under analogous conditions: a kernel size of $7\times7$ and a training duration of 50 epochs. The performance with the Kaiming initialization stood at 76.06\%. In contrast, our innovative initialization method yielded a superior result, registering an accuracy of 76.40\%.

Finally, we explored the effectiveness of our method with larger kernel sizes using the ConvMixer-512 model. Specifically, we employed a kernel size of 9 in the ConvMixer architecture. This approach resulted in a significant improvement, with an accuracy increase of 2.34\%, thereby affirming even better efficacy of our method when applied to models with larger kernel configurations.



\begin{figure}[h]
  \centering
  \begin{subfigure}{0.24\linewidth}
    \includegraphics[width=\linewidth]{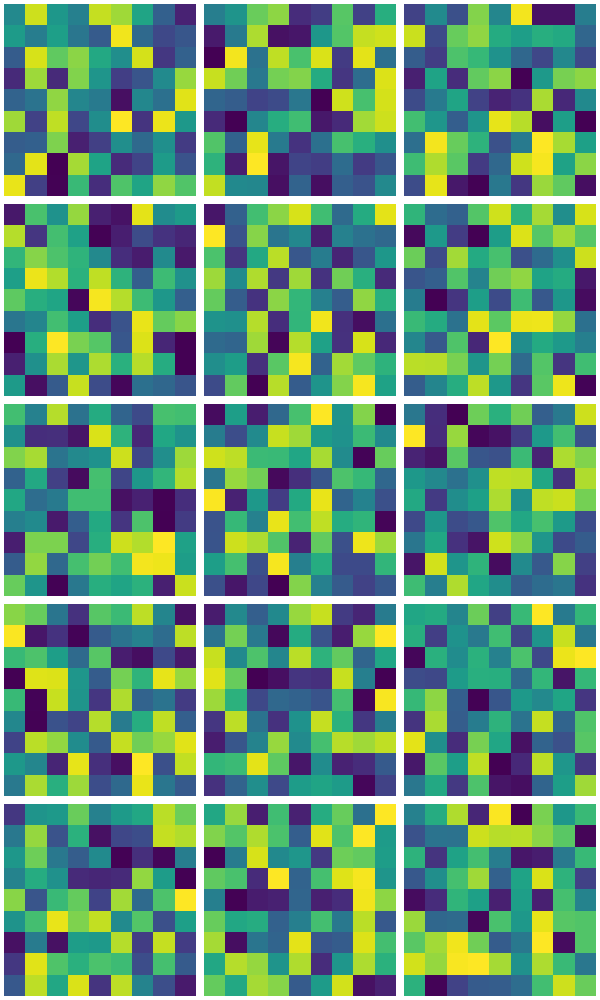}
  \end{subfigure}
    \hfill
  \begin{subfigure}{0.24\linewidth}
    \includegraphics[width=\linewidth]{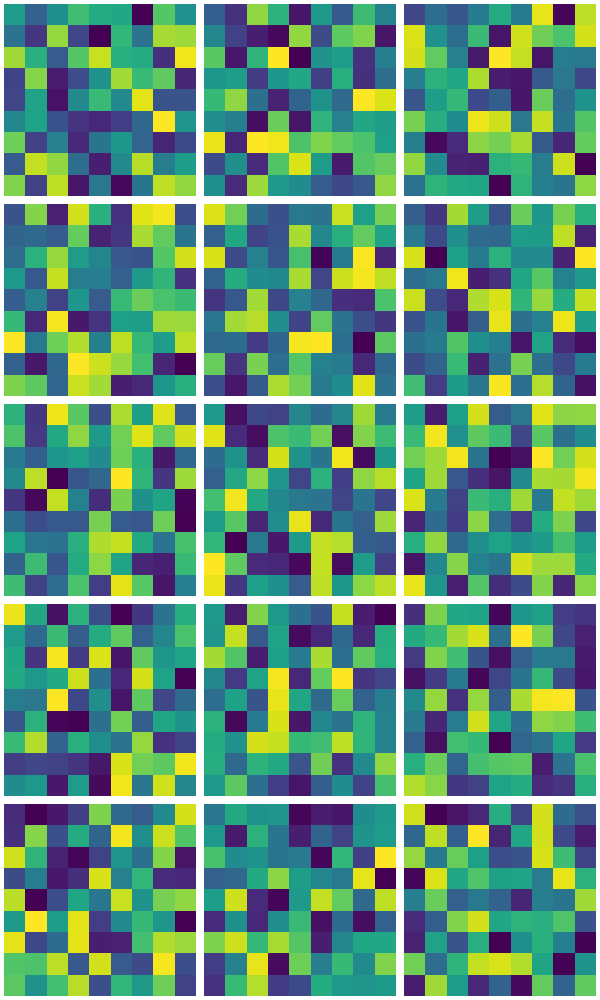}
  \end{subfigure}
    \hfill
  \begin{subfigure}{0.24\linewidth}
    \includegraphics[width=\linewidth]{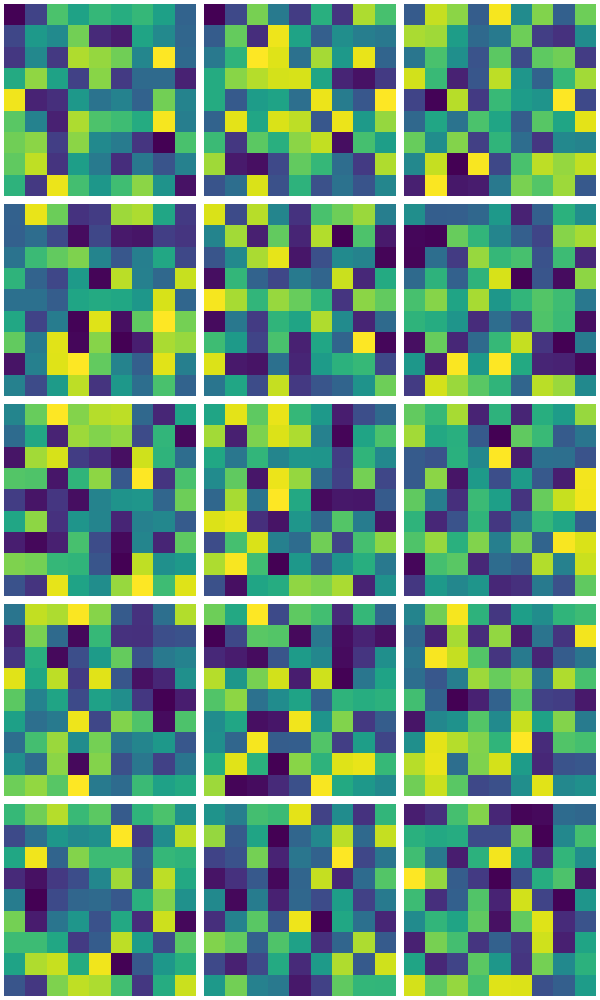}
  \end{subfigure}
    \hfill
  \begin{subfigure}{0.131\linewidth}
    \includegraphics[width=\linewidth]{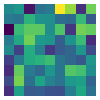}
    \includegraphics[width=\linewidth]{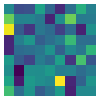}
    \includegraphics[width=\linewidth]{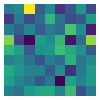}
  \end{subfigure}

   \caption{Clusters from kernels trained on the Cifar10 dataset (left) and the average of each cluster (right).}
   \label{fig:cifar10kernels}
\end{figure}

\vspace*{-0.1ex}\textbf{Cifar10.} We evaluated our initialization on Cifar10 with models with kernel sizes of 5, 7, and 9. However, across different runs, we observed little to no improvements. This may be primarily attributed to the lack of discernible patterns in filters trained on Cifar10, in contrast to their ImageNet counterparts. This disparity is likely rooted in the significant size difference between the datasets, both in number of classes and image sizes. Clusters of kernels trained on Cifar10 are depicted in Figure~\ref{fig:cifar10kernels}.


\subsection{Ablation Study}

To assess our initialization's impact, we conducted an ablation study on the ConvMixer model, as this model exhibited the most significant improvement from our initialization method. The study, performed on the ImageNet dataset, is detailed in Table \ref{tb:ablation}.


Our baseline evaluation using Kaiming initialization achieved 64.00\% accuracy. We applied our initialization, adjusting the DoG function parameters and the excitatory/inhibitory center arrangement. First, we sampled $\gamma$ from a uniform distribution between (0,1) and used both On (excitatory) and Off (inhibitory) centers. This yielded an improved accuracy of 65.20\%.

Next, we narrowed the range of $\gamma$ to (0,0.5), while only utilizing On centers. The resulting accuracy, though slightly lower at 64.78 percent, still exceeded the baseline Kaiming initialization. This suggests that the selection of $\gamma$ and center types both play significant roles in the performance.

Finally, maintaining $\gamma$ in the (0,0.5) range, we reintroduced both On and Off centers into our model. This resulted in the highest observed accuracy of 66.34\%. It is clear from this ablation study, that the selection of $\gamma$ and the type of centers (On or Off) significantly influence the model performance.

\begin{table}
\small
  \centering
  \caption{Ablation on initialization settings with ConvMixer-512 with kernel size $9\times9$ on ImageNet.}
 
\begin{tabular}{lc}
    \toprule
    \textbf{Initialization}  &  Accuracy\\     
    \midrule
    Kaiming & 64.00\\
    Ours, $\gamma \in$  (0,1), On and Off Centeres & 65.20\\
    Ours, $\gamma \in$ (0,0.5), Only On Centers & 64.78\\
     Ours, $\gamma \in$ (0,0.5), On and Off Centeres & \textbf{66.34}\\
    
    \bottomrule
  \end{tabular}
  \label{tb:ablation}
  \vspace{-2.5ex}
  
\end{table}

\section{Conclusions and Future Work}
\textbf{Conclusions.} This paper has delved into an intriguing discovery of center-surround patterns in depthwise convolutional kernels, highlighting a fascinating interplay between artificial neural networks and natural vision systems. We capitalized on this finding by introducing a novel initialization strategy for depthwise kernels, incorporating the principles of the DoG method, typically utilized in bio-inspired vision models. This unique approach taps into the center-surround antagonism property of retinal ganglion cells, offering enhanced contrast sensitivity, mirroring the proficiency of biological vision systems.

The empirical evidence from our extensive experiments on ImageNet firmly backs the efficacy of our proposed method. Compared to the widely used Kaiming initialization, our technique demonstrated a notable improvement in the accuracy of the models. As illustrated in Figure~\ref{fig:cifar10kernels},
it is also interesting to observe that on the Cifar10 dataset, models do not seem to be able to learn biological kernels so effectively. In our opinion, this is a result of the small size of their images ($32{\times}32$ compared to $224{\times}224$ in ImageNet), and the very limited number of their classes (10 compared to 1000 in ImageNet).

\textbf{Future Work.} While the results obtained are promising, there's still scope for further exploration. 
A promising direction is to test our initialization method across a broader range of CNNs, potentially advancing a ubiquitous biology-inspired initialization approach. 

Furthermore, the primary aim of this paper was to highlight the resemblance between trained kernels and their biological counterparts. We have not yet embarked on any form of hyperparameter search to fine-tune the parameters of our initialization method. Parameters like the range for $\gamma$, the proportion of excitatory and inhibitory kernels, initialization specific to each layer, and the balance of positive and negative values in the Difference of Gaussians function have been left unexplored. Potentially, these factors could be tweaked for optimal performance. 

Finally, this paper has not explored the patterns in the "Others" cluster of Figure~\ref{fig:histograms}. It is very likely that these patterns are linked to biological neural processing, too.
\vspace{-1.1ex}
\section{Acknowledgements}
Z.B. is supported by the Doctoral College Resilient Embedded Systems, which is run jointly by the TU Wien's Faculty of Informatics and the UAS Technikum Wien.
\newpage
{\small
\bibliographystyle{ieee_fullname}
\bibliography{egbib}
}

\end{document}